\documentclass[review,3p,10pt,times]{elsarticle}

\usepackage{textcomp, gensymb}
\usepackage{amssymb,amsmath,amsfonts,mathrsfs,bm,gensymb}
\usepackage{graphicx,float,subfig}
\graphicspath{ {./figures/} }
\usepackage{line no}

\usepackage{geometry}
\usepackage{hyperref}
\usepackage{color}
\usepackage[section]{placeins}
\usepackage{soul}
\usepackage{ulem}
\usepackage{algorithm}
\usepackage{algpseudocode}
\usepackage{cleveref}
\usepackage{caption}
\usepackage{subcaption}
\usepackage{amsmath}
\usepackage{booktabs}
\usepackage{array}
\usepackage{amsmath, amssymb, bm}
\usepackage{tikz}
\usetikzlibrary{arrows.meta}
\usetikzlibrary{positioning}
\usepackage{setspace}
\usepackage{fontawesome5}

\newcommand{\Gs}{\sigma}
\newcommand{\Gl}{\lambda}

\newcommand{\mfB}{\mathfrak{B}}
\newcommand{\mfG}{\mathfrak{G}}

\DeclareMathOperator*{\argmax}{argmax}

\definecolor{darkgreen}{rgb}{0.0,0.8,0.0}

\crefname{lstlisting}{listing}{listings}
\Crefname{lstlisting}{Listing}{Listings}

\usepackage{listings}
\usepackage{xcolor}
\usepackage{algorithm}
\usepackage{algpseudocode}

\definecolor{codegreen}{rgb}{0,0.6,0}
\definecolor{codegray}{rgb}{0.5,0.5,0.5}
\definecolor{codepurple}{rgb}{0.58,0,0.82}
\definecolor{backcolour}{rgb}{0.95,0.95,0.92}

\lstdefinestyle{mystyle}{
    backgroundcolor=\color{backcolour},   
    commentstyle=\color{codegreen},
    keywordstyle=\color{magenta},
    numberstyle=\tiny\color{codegray},
    stringstyle=\color{codepurple},
    basicstyle=\ttfamily\footnotesize,
    breakatwhitespace=false,         
    breaklines=true,                 
    captionpos=b,                    
    keepspaces=true,                 
    numbers=left,                    
    numbersep=5pt,                  
    showspaces=false,                
    showstringspaces=false,
    showtabs=false,                  
    tabsize=2
}

\hypersetup{
    colorlinks=true,
    linkcolor=black,
    citecolor=black,
    filecolor=black,
    urlcolor=black,
    pdfauthor=author
}

\makeatletter
\def\ps@pprintTitle{%
	\let\@oddhead\@empty
	\let\@evenhead\@empty
	\def\@oddfoot{}%
	\let\@evenfoot\@oddfoot}
\makeatother

\begin{document}
\begin{frontmatter}

\title{Text to model via SysML: Automated generation of dynamical system computational models from unstructured natural language text via enhanced System Modeling Language diagrams}
 
\author[add2]{Matthew Anderson Hendricks}
\author[add2]{Alice Cicirello \corref{corr1}}

\cortext[corr1]{Corresponding authors}
\address[add2]{Department of Engineering, University of Cambridge, Trumpington Street, Cambridge, UK}
 
\begin{abstract}

This paper contributes to speeding up the design and deployment of engineering dynamical systems by proposing a strategy for exploiting domain and expert knowledge for the automated generation of a dynamical system computational model starting from a corpus of documents relevant to the dynamical system of interest and an input document describing the specific system. This strategy is implemented in five steps and, crucially, it uses system modeling language diagrams (SysML) to extract accurate information about the dependencies, attributes, and operations of components. Natural Language Processing (NLP) strategies and Large Language Models (LLMs) are employed in specific tasks to improve intermediate outputs of the SySML diagrams automated generation, such as: list of key nouns; list of extracted relationships; list of key phrases and key relationships; block attribute values; block relationships; and BDD diagram generation.  The applicability of automated SysML diagram generation is illustrated with different case studies. The computational models of complex dynamical systems from SysML diagrams are then obtained via code generation and computational model generation steps. In the code generation step,  NLP strategies are used for summarization, while LLMs are used for validation only.
The proposed approach is not limited to a specific system, domain, or computational software. Domain and expert knowledge is integrated by providing a set of equation implementation templates. Specific examples are used to compare individual steps of the proposed approach with LLM generation: keyword extraction and code generation based on the project structure. In all cases, the proposed approach demonstrates a quality that is closer to ground truth than the LLM counterpart. This work represents one of the first attempts to build an automatic pipeline for this area.
The applicability of the proposed approach is shown via an end-to-end example from text to model of a simple pendulum, showing improved performance compared to results yielded by LLMs only in zero-shot mode.
\end{abstract}

 
\end{frontmatter} 

\section{Introduction}

In many engineering and science applications, it is important to analyze and predict the wide range of behaviors of real world systems over time for a broad range of inputs \cite{Schouken,SystemIDbook}. Therefore, a physical system (e.g., a car) is often ``virtually tested" by using a dynamical system computational model to reduce engineering cycles and avoid expensive modifications of the manufactured system. Systems are objects that for some given signals (inputs and disturbances) and initial conditions produce an output signal (e.g., vibration response) ~\cite{SystemIDbook}. These dynamical systems are  typically composed by many components, which may interact with each other, and for which a linear time-invariant modeling assumption is often inadequate \cite{Schouken}. Depending on the application, multi-scale, multi-physics, high-dimensional, strongly nonlinear, and time-dependent models might be required \cite{Schouken}. Since the model would be representative of only certain aspects of the real world, the fidelity and complexity of the model would depend on the task for which that model is going to be used.  Creating computational models for complex dynamical systems at the initial design stages is often a time-consuming and error-prone process that requires significant expertise and resources since:  (i) many components of a system are used to satisfy simultaneously different performance constraints and tasks (e.g., the suspension of a car is important for noise and vibration harshness performance, ride performance, handling performance), therefore affecting the modeling strategy and decision-making of different design teams; (ii)
information on components dependencies are available in a broad range of textual resources (e.g., specifications  \cite{dori2004smart}, manuals, technical reports, maintenance reports) and without access to data from the real world (since no prototype design is available); (iii) heavily rely on engineers’ expertise and professional judgment, leading to lack of completeness or exhaustiveness of the analysis, inconsistencies across models fidelity, as well as an implicit degree of subjectivity.
As a result, currently, ensuring a comprehensive analysis throughout the design phase is challenging and typically requires multiple costly cycles of design, prototyping and testing. This lengthy iterative process can introduce inconsistencies across design decisions, models, and documentation, as well as a significant degree of subjectivity \cite{sawyer2005shallow,arellano2015frameworks, zhong2023natural}. Ultimately, this might lead to costly modifications of the manufactured product \cite{arellano2015frameworks}.

Recently, Large Language Models (LLMs) have received particular attention for code completion, synthesis, analysis, and interpretation of the results, and by converting natural language text into domain-specific languages for interaction with software tools \cite{brown2020languagemodelsfewshotlearners,nijkamp2023codegenopenlargelanguage, wang2023codet5opencodelarge, Roziere2023_CodeLlama, rupprecht2025text2model}. The use of LLMs for generating dynamic models of chemical reaction systems from descriptions written in natural language has recently been explored \cite{rupprecht2025text2model}. Specifically, the Llama 3.1- 8B LLM model \cite{LLAMA} was fine-tuned by using low-rank adaptation on synthetically generated Modelica code for different reactor scenarios to increase the semantic and syntactic accuracy of automatically generated domain-specific programming language instructions in Modelica \cite{rupprecht2025text2model}.   However, these LLMs-based approaches are often limited in their ability to generalise to unseen dynamical system simulations scenarios that were not included in the training data, as demonstrated and discussed in \cite{rupprecht2025text2model}. 
Moreover, these approaches do not provide a structured representation of the system, which is crucial for understanding the interaction between components.

A strategy is proposed for  automatically generating  computational models of the dynamical system directly from unstructured natural language text by exploiting System modeling  Language (SysML) diagrams, that are a commonly used tool in systems engineering \cite{kossiakoff2011systems,friedenthal14practical,hart2015introduction,delligatti2013sysml,huang2007system}, while  making strategic use of Natural Language Processing (NLP) strategies \cite{jurafsky2009speech} and limited use of LLMs. 
SysML is a subset of the Unified modeling Language (UML), and specifically it is graphical modeling language  used in systems engineering \cite{friedenthal14practical,delligatti2013sysml,huang2007system,omg19omg,friedenthal08omg,hause06thesysml}. Recently, SysML has gained importance  as a critical enabler of Model-based Systems Engineering \cite{friedenthal14practical,hart2015introduction, zhong2023natural, SANTOS2025112407}. 
The basic elements of SysML diagrams include blocks and their relationships. A block is an elemental modeling construct in SysML that represents both real entities, such as physical objects, and abstract entities, such as concepts~\cite{friedenthal14practical}. Among the most descriptive SysML diagrams for system engineering, the Block Definition Diagram (BDD) is the most useful type of  diagram for building a computational model of a dynamical system. The BDD describes the system components and their attributes (properties, behaviors, constraints) in terms of blocks, and their relationships by means of connecting lines.  Typically they are used for higher level descriptions, but not for direct modeling generation \cite{SANTOS2025112407}. 

The proposed approach does not intend to replace engineering personnel. Instead, it is intended to be used to assist engineers in the  generation of computational models for dynamical systems from unstructured natural language text by exploiting domain and expert knowledge. The proposed approach improves performance with respect to LLMs, and it yields intermediate outputs that can be controlled by the engineering team. Consequently, it would enable faster investigations of the dynamical system performance by investigating  simulations of the system under various operational and environmental conditions, including extreme events.  The proposed approach is designed to be general and applicable to a wide range of dynamical systems, allowing for the generation of computational models from unstructured natural language text only. The approach is not limited to any specific type of system or domain (i.e., it is not context specific), making it versatile and adaptable to various engineering contexts and in other scientific fields. Moreover, each step in this approach is not implementation-specific, meaning that each  step could be carried out by exploiting alternative tools or algorithms that produce the same outputs for the set of given inputs at a step. This allows future implementations of the approach to be carried out using improved tools that outperform those that exist today or are more suited to the task at hand. The key enabler of the proposed approach is exploiting and enhancing strategies for automatically generating system representations in the form of SysML diagrams \cite{zhong2023natural, SANTOS2025112407} as an intermediate step towards generating computational models.
The code generation step converts the SysML diagram into code that models the current dynamical system behavior and can be modified by the user for further investigations. Domain and expert knowledge on the specific dynamical system is integrated in this step by including manually compiled templates of equation implementation. Finally, the computational model generation step uses the generated code to run simulations of the dynamical system across various input conditions.

To the best of the authors’ knowledge, this is the first study that focuses on the automatic generation of computational models of dynamical systems from unstructured natural language text via SysML diagrams.

The main contributions of this paper are:
\begin{itemize}
   \item An automated approach to generate dynamical system computational models from natural language text by enhancing SysML. 
   \item The enhancement of a context-free automated generation of SysML diagrams by  handling spelling errors, coreferences, and automatically extracting attribute values
   \item Similarity and match scores for comparing extracted and ground truth blocks/attributes, enabling systematic evaluation of SysML diagrams.
   \item A general procedure for building  models of complex dynamical systems from SysML diagrams based on automated code generation and computational model generation.
\end{itemize}

Details on the proposed approach, validation of intermediate steps with respect to manually generated results and results obtained with LLMs, and on and end-to-end case study, are provided in what follows. The results show that the proposed approach is competitive against zero-shot prompting of LLMs in the extraction of key phrases and the generation of BDD diagrams.

\section{Automated generation of dynamical system computational models from natural language text by enhancing SysML }\label{sec:autogenDynamicalFromSysML}

\begin{figure}[H]
		\centering
        \begingroup
        \setstretch{1}
		\begin{tikzpicture}[node distance=3.5cm, every node/.style={align=center}]
			\node (input) [rectangle, text width=2.5cm, draw] {Input: Unstructured Natural Language Text};
			\node (prep) [rectangle, text width=3cm, draw, below of=input, node distance=2cm] {Text\\ Preparation\\and Processing};
			\node (kg) [rectangle, text width=2.5cm, draw, right of=prep, node distance=3.5cm] {Knowledge Graph\\Generation};
			\node (sysml) [rectangle, text width=2.5cm, draw, fill=gray!20, right of=kg, node distance=3.5cm] {\textbf{SysML\\ Diagram\\Generation}};
			\node (code) [rectangle, text width=2.5cm, draw, right of=sysml, node distance=3.5cm] {Code\\Generation};
			\node (sim) [rectangle, text width=2.5cm, draw, right of=code, node distance=3.5cm] {Computational \\model \\
            generation};
			\node (output) [rectangle, text width=2.5cm, draw, above of=sim, node distance=2cm] {Output:\\Simulation Result};
			\draw[->, thick] (input) -- (prep);
			\draw[->, thick] (prep) -- (kg);
			\draw[->, thick] (kg) -- (sysml);
			\draw[->, thick] (sysml) -- (code);
			\draw[->, thick] (code) -- (sim);
			\draw[->, thick] (sim) -- (output);
		\end{tikzpicture}
        \endgroup
		\caption{Overview of the steps for automatic generation of dynamical systems computational models from unstructured natural language text.}
		\label{fig:overview_flowchart}
	\end{figure}
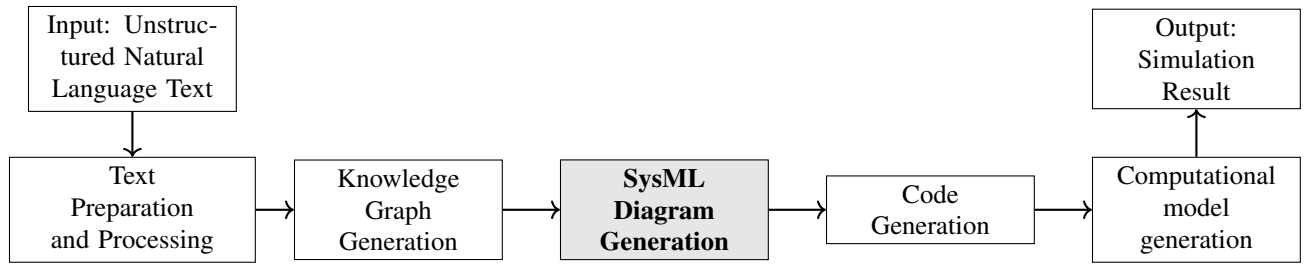

 The proposed approach enables the automated generation of computational models to investigate the performance of a dynamical system starting from domain and expert knowledge in the form of unstructured natural language text describing the dynamical system of interest by exploiting System modeling  Language (SysML) diagrams, and in the form of a set of code templates. It consists of the steps summarised in \Cref{fig:overview_flowchart}.

	\begin{enumerate}
		\item \textbf{Text Preparation and Preprocessing}: A manually selected corpus of text document with no specific structure or writing style that is relevant to the dynamical system of interest is prepared and preprocessed. Only natural language text in the main text of each document is retained. By using standard Natural Language Processing (NLP) techniques \cite{jurafsky2009speech}, the text is preprocessed by removing stop words and punctuation, converting casing to lowercase, tokenizing and lemmatizing words. In addition to these steps, spelling correction and coreference resolution strategies are also implemented. 
		\item \textbf{Knowledge Graph Generation}: The preprocessed text is converted into a knowledge graph, that is a representation of the relationships between the components of a system \cite{fensel2020introduction}, by exploiting general NLP strategies. 
		\item \textbf{SysML diagrams Generation}:  The  Block Definition Diagram (BDD) diagram is a SysML diagram that represents engineering systems by showcasing components alongside their attributes and operations. The knowledge graph is converted into a BDD diagram by improving upon the approach proposed in \cite{zhong2023natural} with the inclusion of a strategy for attribute value extraction. 
		\item \textbf{Code Generation}: The BDD diagram is used to generate code that models each component of the dynamical system behavior and can be altered for further experimenting. This step is divided into 3 substeps:
		\begin{enumerate}
            \item \textbf{File Structure Generation}: an hierarchical structure of folders with blank files is automatically generated from the BDD to mimic the hierarchy of the dynamical system' components. 
			\item \textbf{Class Skeleton and Code Comment Generation}:  The BDD diagram is the mapped deterministically into a code skeleton of each component with code comments to provide the foundation of the computational model. Specifically, each  blank file generated in the previous step, is automatically populated with a class that represents the component,  with code comments providing descriptions of the functionality of each component.
			\item \textbf{Function Body Generation}: The code skeleton and code comments are used to implement the function bodies with the relevant equations of motion in the class skeleton. Within this step, a set of equation implementation templates is provided by engineers with domain knowledge on the system. 
		\end{enumerate}
		\item \textbf{Computational model generation}: 
        Each file in the hierarchical structure now contains the information to model the system. A top-level system file is generated here that initializes (i.e., creating an instance of a class) all the top-level components of the system. Using the initial conditions and simulation parameters (e.g. the length of the simulation) manually specified by the user, a simulation is run for all the classes in the system with functions containing the implementation of the equations of motion and the resulting states at each timestep form the output of the simulation.
	\end{enumerate}

The proposed strategy for context-free automated generation of SysML diagrams from unstructured natural language text, comprising steps 1 to 3 is detailed in \Cref{sec:SysMLg}. The steps for implementation and validation of these strategies are discussed in \Cref{sec:ImplAndValidModels}. 
The steps necessary to transform the SysML diagram into a computational model (steps 4-5) are discussed in \Cref{sec:Sysmltomodel}. Finally, and end-to-end validation case study for a dynamical system is investigated in \Cref{sec:validation}. 

\section{Context-free automated generation of SysML diagrams from unstructured natural language text}\label{sec:SysMLg}

The procedure for automatic generation of the SysML diagram from unstructured natural language text. The approach is not tuned to a specific system or domain (i.e., it is context-free), and it builds on the strategy proposed in \cite{zhong2023natural}. It consists of the following steps: 

\subsection{Text preparation and pre-processing}\label{sec:prep}

The unimodal data source used within the proposed approach is unstructured natural language text without a predefined writing style, describing the dynamical system for which a computational model is going to be automatically built.  A corpus of text documents describing the class of dynamical systems of interest  is manually selected and used to automatically extract information to obtain a computational model of a specific dynamical system of interest to the user.  It is worth emphasizing here that information contained in tables, equations, and figures included in each document is discarded, and only textual information is retained.  Based on the results obtained in \cite{zhong2023natural} for automated generation of SysML, the size of the corpus could be considered sufficient when the document count is larger than 100, and the average word count per document is larger than 500. Typically, this size of corpus is obtained by splitting large documents (e.g., books, manuals, reports, patents, standards) into chapters and sections to achieve a sufficient document count \cite{zhong2023natural}. However, when dealing with documentation referring to a specific dynamical system, it might not be possible to have such large corpus of documents. Therefore, in the present work a range of examples is considered. In the first case,  the corpus consist of the chapters of the book  \cite{jackman1912flying}, and therefore comprises less than 30 documents. In the second case, a corpus of patents with only 10 documents is considered. The third example, that is used as an end-to-end case study  in \Cref{sec:validation} consider a document under 500 words and uses the same corpus of patents. A summary of the input documents and corpuses used in the paper is shown in \Cref{tab:document_types}.

Specific strategies for handling spelling errors and coreferencing, are detailed detailed in what follows. 

\subsubsection{Handling spelling errors}\label{sec:sperr}
Spelling errors are common in human-generated documents, and for example, in the \cite{jackman1912flying} corpus the word ``ornithopter" was misspelled as ``ornithoper" as shown in \Cref{fig:spell_error_example}. Spelling errors might cause misidentification of system components or attributes that are necessary for the computational model generation. Approaches for spelling corrections can be classified into: rule-based, statistical models (based on N-grams or on edit distance) and deep learning (e.g., transformers) approaches \cite{hladek2020survey}. Given the component of a dynamical system would be described with specialised words, for which the corpus might be relatively limited, statistical models based on edit distance are chosen. 

\begin{figure}[H]
		\centering
		\fbox{
			\begin{minipage}{0.8\textwidth}
				\centering
				``Of these the aeroplane takes precedence and is used almost exclusively by successful aviators, the helicopters and \textcolor{red}{ornithopers} having been tried and found lacking in some vital features, while at the same time in some respects the helicopter has advantages not found in the aeroplane."
			\end{minipage}
		}
		\caption{A excerpt from the book \cite{jackman1912flying} with the word ``ornithopter" misspelled as ``ornithoper".}
		\label{fig:spell_error_example}
	\end{figure}

In this work, the spelling correction process is carried out using a simple yet effective probabilistic approach based on edit distance, which exploits  Bayes' Theorem. Specifically, the goal is to identify out of the possible candidate corrections $c$ (that belongs to a set of candidates $C$, $c \in C$), the one $\hat c$ that maximizes the probability that the correction  is the intended correction given an original misspelled word $w$. In probability terms ($P ( \cdot) $) this can be written as \cite{norvig2009natural}: 
	\begin{align}
		\hat c=\arg \max_{{c \in C}}  P(c|w) =  \arg \max_{{c \in C}} \frac{P(w|c)P(c)}{P(w)} \label{eq:bayes}
	\end{align}

	The two significant parts of this method are the language model $P(c)$ and the error model $P(w|c)$, since $P(w)$ is the same for all the possible $c\in C$ of the word $w$. The language model is typically formed by considering English Language text which it is believed to be free from spelling errors.  This text can be obtained by concatenating public domain book excerpts and wikipedia documents.  The $P(c)$ is then obtained by counting the number of times that the candidate correct word appears in chosen text file, over the total number of words in the language model. It was found that combining standard text with text containing specialised words referring to the dynamical system of interest (e.g., internal documentation from the engineering team) is key in engineering applications. For example, engineering documents often include domain specific words such as the word ``ornithopter" are found in the used corpus in \cite{jackman1912flying}. 
    
    The error model $P(w|c)$ can be defined by assigning a probability of 1 if the word belongs to the known words. Alternatively, equal probabilities are assigned to candidate words that have the same edit distance from the word being corrected. The edit distance is the number of operations needed to convert one word into another and is defined as the minimum number of insertions, deletions or substitutions needed to convert one word into another. An edit distance equal to 1 (Edits$_1$), would correspond to remove one letter (e.g., deletion), to swap two adjacent letters (e.g., transposition), to changing one letter to another (e.g., replacement)  or to adding a letter (an insertion). Similarly, (Edits$_2$) would consider and edit distance equal to 2.  These edits can be automatically implemented for each word, to return all edited strings that can be made with a simple edit or with two simple edits for a word \cite{norvig2009natural}, and only up to edit ditstance 2 are going to be considered. If none of these three conditions are met, the $P(w|c)$ is going to be set to zero. This error model can be expressed as:

	\begin{align}
		P(w|c)=
		\begin{cases}
		1 & \text{if } w \in \text{KnownWords} \\
		\displaystyle \frac{1}{|\text{Known}(\text{Edits}_1(w))|} & \text{else if } \text{Known}(\text{Edits}_1(w)) \neq \emptyset \\
		\displaystyle\frac{1}{|\text{Known}(\text{Edits}_2(w))|} & \text{else if }  \text{Known}(\text{Edits}_2(w)) \neq \emptyset \\
		0 & \text{otherwise}
		\end{cases}
	\end{align}

    For cases where $P(c|w) = 0$, the word will not be corrected. For the use cases investigated in this paper, it was found that most simple spelling errors were corrected. Moreover, when spelling errors on domain specific words were present, it was found that those were successfully corrected if the correct spelling of the domain specific words were included into the language model. The spelling correction has been carried out by using the filyp/autocorrect package \cite{pyautocorrect}.
    It is worth mentioning that alternative strategies for spelling error detection based on Sequence-to-Sequence (Seq2Seq) models \cite{sun2020seq2seqtranslationmodelsequential} architecture can take into account the overall context of the text, and therefore could lead to results improvement.

\subsubsection{Handling Coreference }\label{sec:coref}
Coreference is the phenomenon when two or more different phrases within a text refer to the same entity \cite{hobbs1979coherence}.  Coreference resolution is the process of identifying ``coreference chains" and listing all words that refer to the same entity. Typically, this is considered an easy task for a human, but it requires careful attention to be automated. This step is critical for the automatic generation of knowledge graphs and  SysML diagrams, since they rely on the identification of key nouns and key phrases in the text. If the identified noun or phrase refers to the wrong entity, it won't be possible to properly categorize the correct information together. This is especially detrimental in the case of the repeated use of pronouns or other implicit mentions, like ``it" and ``machine", as these words can be used to refer to different entities in the text. A system without coreference resolution would lump the information under the  entities ``it" and ``machine". Consequently, this would result in a loss of information and a misrepresentation of the system in the knowledge graph and SysML diagram. 

For example in \Cref{fig:coref_resolution_example}, all the implicit mentions of the noun ``screw propeller" with the word ``its" has been made explicit through coreference resolution to the noun ``propeller".  Without coreference resolution in \Cref{fig:coref_resolution_example}, the information on the ``screw propeller's" maximum efficiency, load and pitch speed would not have been extracted into the knowledge graph.

     In this paper, this task is carried out by using the coreference tool developed in \cite{hudson_coreferee}, called ``Coreferee tool''. This tool uses a mixture of neural networks and programmed rules to identify coreference chains in the text \cite{hudson_coreferee}. Since the focus is engineering documents describing dynamical systems, it is plausible to assume that the first mention of an entity is the most explicit one. After identifying these coreference chains, all the instances referred to the same entity are replaced with the word used in its first mention. 
	
It is worth noting that the word ``screw propeller" is replaced by ``propeller". This is because the tool used tends to prefer single word replacements more than longer phrases. The development of enhanced coreference tools is considered beyond the scope of this work. 

\subsubsection{Natural Language Processing (NLP) Preprocessing Steps}\label{sec:nlpProc}
NLP preprocessing steps are applied to the entire corpus with the goal to support noun extraction \cite{jurafsky2009speech}.  The NLP preprocessing steps considered are the same as those described in \cite{zhong2023natural}. They consist of four steps that are implemented by using the Natural Language Toolkit (NLTK) \cite{bird2009natural}. To clarify each step, the example sentence ``A screw propeller working under load approaches more closely to its maximum efficiency." is going to be used.

	\begin{enumerate}
		\item \textbf{Tokenization}: Natural language text is broken down into smaller units called tokens, typically referring to words \cite{jurafsky2009speech}. In this study, both sentence tokenization (using the adapted Punkt sentence augmenter\cite{kiss2006unsupervised}) and word tokenization (using the Penn Treebank Tokenizer \cite{bird2009natural}) were carried out on each document. The tokenization strategy implemented only uses one-word unigrams. \\
		\textit{Example Result with each token seprated by a comma:} [A, screw, propeller, working, under, load, approaches, more, closely, to, its, maximum, efficiency, .]

		\item \textbf{Part-of-Speech (PoS) Tagging}: Each word is then classified into different lexical categories based on its role within the sentence \cite{jurafsky2009speech}. This is done by using the Greedy Averaged Perceptron tagger along with the Penn Treebank tagset \cite{bird2009natural}. This allows us to filter and retain nouns for further processing steps. \\
		\textit{Example Result:} [(A, DT), (screw, NN), (propeller, NN), (working, VBG), (under, IN), (load, NN), (approaches, VBZ), (more, RBR), (closely, RB), (to, TO), (its, PRP), (maximum, JJ), (efficiency, NN), (., .)] where the tags (e.g. DT and NN) and their meanings are shown in \Cref{tab:ptb-selected-tags}.

		\item \textbf{Lemmatization}: The WordNet Lemmatiser is used to convert the nouns into lowercase and reduce them to their root forms \cite{bird2009natural}. \\
		\textit{Example Result:} [a, screw, propeller, work, under, load, approach, more, closely, to, its, maximum, efficiency,.]

		\item \textbf{Stop Word Removal}: Stop words are words frequently occurring and have low significance in for the task of extracting information from text. This list depends on the language being used and also on the context of the documents.  The NLTK's list of English stop words \cite{bird2009natural} is used to identify those words and remove them. \\
		\textit{Example Result:} [screw, propeller, work, load, approach, closely, maximum, efficiency], where [a, under, to, its] were removed.
	\end{enumerate}

It is worth mentioning that alternative approaches for lemmatization and stopword removal can be implemented by using advanced language models, such as the language model BERT (Bidirectional Encoder Representations from Transformers) \cite{devlin2019bertpretrainingdeepbidirectional}. Such models would take into account the overall context in text by reading in both directions simultaneously.

\subsection{Knowledge Graph Generation}\label{sec:Kgrap}
The knowledge graph generation process is important to allow the proposed algorithm to identify the interactions within the dynamical system before they are augmented into diagrams and code. The knowledge graph generation process is built according to the methodology proposed in \cite{zhong2023natural} and consists of the following steps:

	\begin{enumerate}
		\item \textbf{Key Noun Extraction}: Key nouns are extracted from the preprocessed text using the term frequency - inverse document frequency (tf-idf) metric.
		\item \textbf{Relationship Extraction}: Relationships between the nouns of every sentence is extracted using the OpenIE toolbox \cite{openIE5}.
		\item \textbf{Key Phrase and Key Relationship Selection}: Key phrases and key relationships are selected by using ``importance metrics''.
	\end{enumerate}

	\subsubsection{Key Noun Extraction}
	\label{sec:key_noun_extraction}

A key noun is a noun that is likely to be in the phrase representing any significant component in the dynamical system extracted. Therefore, 	the  extraction of key noun is at the core for the identification of the components of the system.  An example is shown in \Cref{fig:key_noun_example}.

	\begin{figure}[H]
		\centering
		\fbox{
			\begin{minipage}{0.8\textwidth}
				\centering
					``''\textcolor{darkgreen}{Gliders} as a rule have only one \textcolor{darkgreen}{rudder}, and this is in the rear. It tends to keep the apparatus with its head to the wind. Unlike the \textcolor{darkgreen}{rudder} on a boat it is fixed and immovable. The real \textcolor{darkgreen}{motor-propelled flying machine}, generally has both \textcolor{darkgreen}{front} and \textcolor{darkgreen}{rear} \textcolor{darkgreen}{rudders} manipulated by \textcolor{darkgreen}{wire cables} at the will of the \textcolor{darkgreen}{operator}."
			\end{minipage}
		}
		\caption{A excerpt from the book \cite{jackman1912flying} with  key nouns highlighted in green.}
		\label{fig:key_noun_example}
	\end{figure}

 Key nouns can be automatically identified by using the tf-idf score, since this selects words that occur frequently in the selected document but filters out those that appear frequently throughout the selected corpus. Typically this score is applied after preprocessing of the documents, so that they contain only nouns. 
 
The tf-idf score is calculated with the equation shown in \Cref{eq:tfidf} where $f_{w,d}$ is the raw count of word $w$ in the document $d$, $N_d$ is the number of words in the document $d$, $N_c$ is the number of documents in the corpus $c$, and $n_{w,c}$ is the number of documents in the corpus $c$ where the word $w$ appears:
	\begin{align}
		\text{tf-idf}(w, d, c) = \log_{10} \left(\frac{f_{w,d}}{N_d} \right) \times \left( \log_{10}\left(\frac{N}{1 + n_{w,c}}\right) +1 \right) \label{eq:tfidf}
	\end{align}

The tf-idf score are then normalized by the highest tf-idf score. Key nouns are identified as those nouns which have the normalized tf-idf score $\text{tf-idf}_n(w, d, c)$ above a user defined threshold $\Gs_{\text{tf-idf}}$, which is a hyperparameter that can be tuned. 

	\subsubsection{Relationship Extraction} \label{sec:rs_ext}
	A relationship is defined by a triplet of (subject, relation, object) phrases. These relationships can be extracted from each tokenized sentence in the text, an example is shown in \Cref{fig:rel_extraction_example}. The reason the sentences are only tokenized and not fully preprocessed is because the other preprocessing step will interfere with the relationship extraction process. The relationship, and the associated confidence score, are extracted by using the OpenIE Toolbox \cite{openIE5}. These relationships can be filtered by setting a threshold on the confidence score  $\Gs_{\text{relationship}}$, that is user-defined hyperparameter. It is worth mentioning that the OpenIE toolbox extracts relationships by combining four methods: Semantic Role Labelling (SRL) \cite{christensen2011analysis}; Relational Noun Based Information Extraction (IE) \cite{pal-mausam-2016-demonyms}; Numerical IE \cite{saha-etal-2017-bootstrapping}; and Coordination Analyser \cite{pauls2011faster}. 

    It is worth mentioning that the relationship extraction results are dependent on the performance of the previous steps. Alternatively, the relationship extraction can be carried out directly with a LLM-based approaches.

	\subsubsection{Key Phrases Selection} \label{sec:key_phrase_select}

Candidate key phrases are generated by first preprocessing all the subject and object phrases obtained from the relationship extraction step by using the preprocessing techniques described in \Cref{sec:prep}.  	Key phrases are phrases (intended here as word and multiple words describing either the subject or the object from the extracted relationships) that describe significant components in the dynamical system. The candidate key phrases are then filtered by removing all phrases that do not map to any key nouns found in the key noun extraction phase. In all the remaining phrases, all the words that are not key nouns are also removed from the phrase. To prevent phrases that are too long (many words describing either the subject or the object), we define a maximum length for the phrases $L_{\text{phrase}}$ which is a user-defined hyperparameter. For any phrase whose length exceeds $L_{\text{phrase}}$, the words with the lowest tf-idf scores are removed until the phrase is of length $L_{\text{phrase}}$. 

	Finally, key phrases are selected by choosing phrases that had a high number of repetitions throughout the whole document, and whose constituent terms (words) on average have simultaneously a low degree of specificity and a high normalized tf-idf score, as proposed in \cite{zhong2023natural}. Both the average normalized tf-idf score and repetition count are considered as the repetition count accounts for the frequency of the entire phrase, whereas the average normalized tf-idf score is geared towards single words within the multi-word phrase.
    The degree of specificity of nouns is quantified by using WordNet depth \cite{fellbaum1998wordnet}, where a high WordNet depth would refer to something very specific such as the word ``artery", whereas very general words such as ``entity" would have a relatively low WordNet depth. This is derived from WordNet, a large lexical database of English, where the each word's meaning is represented as a cognitive synonym set called synsets. The normalized WordNet depth $h$ of each word is found by dividing by the largest depth value found in the given document. The complement of the WordNet depth $h' = 1- h$ is then used for the WordNet score.
	
	The importance metric $\Gl_{p,d}$ is defined as \cite{zhao2021natural}:

	\begin{align}
		\Gl_{p,d} = \sum_{i=1}^{N_p} \frac{\text{tf-idf}_n(w, d, c)}{N_p} +  \sum_{i=1}^{N_p} \frac{h'_{i,d}}{N_p} + \text{count}_{p,d} \label{eq:importance_metric}
	\end{align}
	
	where $\text{tf-idf}_n(w, d, c)$ is the normalized tf-idf score of the $i$th word in the phrase $p$ in document $d$, $h'_{i,d}$ is the normalized WordNet score of the $i$th word in the phrase $p$, $N_p$ is the number of words in the phrase $p$, and $\text{count}_{p,d}$ is the number of times the phrase $p$ appears in the document $d$. Phrases with importance metrics that exceed a user-defined threshold $\Gs_{p}$ are retained as key phrases.

It is worth mentioning that the WordNet project contains only a limited vocabulary and is no longer being developed, but it can  still can be used as a source of additional information. Alternative approaches such as KeyBERT (Python library for keyword and keyphrase extraction using BERT embedding) \cite{grootendorst2020keybert} would be more robust to vocabulary expansion, the use of synonyms, morphology. 

	\subsubsection{Key Relationship Selection}
	Key relationships are defined as relationships (defined in \Cref{sec:rs_ext}) where both the subject and object phrases are identified as key phrases via the methodology in \Cref{sec:key_phrase_select}. This helps prevent open-ended relationships from cluttering the resulting knowledge graph. An example knowledge graph is shown in \Cref{fig:graph_relationships}.

It is worth mentioning that the Knowledge Graph Construction Pipeline considered is effective when dealing with high-quality, correct texts with a limited vocabulary, making the coreference extraction step (detailed in section 3.1.2) critical. The use of LLM would eliminate the need for separate coreference calculations.

\subsection{Block Definition Diagrams generation}\label{sec:BDDGen}
Although knowledge graphs can capture information about the system and its components, it is not presented in a system engineering form that could facilitate the evaluation of the underlying dynamical system computational model. In system engineering, the most common way to represent the system and its components is through SysML diagrams.  In the paper \cite{zhong2023natural},  three different types of SysML diagrams were generated from natural language text, namely Block Definition Diagrams (BDDs), Internal Block Diagrams and Requirement Diagrams. The BDD diagram provides a high-level representation of the system and its components, and it is used to communicate the design of the system to stakeholders and other engineers. This is the most applicable type of SysML diagram for building a computational model of a dynamical systems, since the system and its components are represent by blocks connected via lines \cite{specification2007omg}. The information in each block includes its label (its name), attributes (i.e., properties, behaviors, constraints), operations (i.e., the different functionalities of the component being represented), and parts (a dynamical component owned by the represented component). It was observed across examples that units such as ``inches" where incorrectly extracted as components and classified as ``parts". These errors can be detected manually by examining the BDD diagram and this can be fixed by adding a filter in the key phrase extraction process in \Cref{sec:key_phrase_select} that compares phrases to a database of well known units and filters out phrases that match those units. Arrows are drawn between separate blocks to indicate categorical relationships between the components.

Since a full BDD diagram would be too complex to be displayed as a figure in this paper, only a sub-diagram of the full BDD diagram from Chapter 7 in the book \cite{jackman1912flying} is shown in \Cref{fig:provided_bdd} by considering the blocks under the ``glider'' block of the full BDD diagram. The labels of the blocks (e.g. ``rudder beam'') identifies the component of the dynamical system the block is representing.  The attribute section of the ``rudder beam block'' shows that the rudder beam component is oriented vertically and is 11 inches and 8 feet long.  For example, the operations section of the rudder beam block demonstrates that the rudder beam component has a ``form'' functionality. Finally, the parts listed in the block enumerates the different components that are owned by the component being represented. For example, the parts section of the frame rudder block showcases that the frame rudder owns the ``rudder beam'' component.

 In the automated generation process of the BDD diagrams proposed in \cite{zhong2023natural}, no strategy for automated attribute extraction was considered. Nonetheless, this step is critical in the present work for identifying the corresponding computational model, since the properties, constraints and behaviors of each component of a dynamical system need to be specified.

The BDD Diagram generation process from a knowledge diagram can be summarised with the following steps:

	\begin{enumerate}
	\item \textbf{Attribute Value extraction}: automated extraction of quantitative and qualitative information on a component from text. This is carried out by using a one shot prompting of a LLM model. 
    \item \textbf{Relationship Mapping}: The relationships between the blocks are mapped to the relationships in the BDD diagram which correspond to lines and arrows. This is done by using the key relationships extracted from the knowledge graph generation process and mapping them to the relationships in the BDD diagram. The mapping is done by using a set of rules that define how the relationships in the knowledge graph correspond to the relationships in the BDD diagram.
		\item \textbf{Block Augmentation}: Some blocks are added to the BDD Diagram to make the graph more complete. The addition of these blocks is achieved by augmenting textual phrases and relationships. These blocks are indicated by dashed lines in the BDD diagram.
		\item \textbf{Diagram Generation}: The BDD diagram is generated by using the relationships and blocks extracted from the knowledge graph.  The BBD diagram is displayed by using an open source tool called PlantUML with Graphviz as the graphical engine \cite{ellson2002graphviz}.
	\end{enumerate}

Steps 1-3 are detailed in what follows. 

\subsubsection{Attribute Value Extraction}\label{sec:attrValExtraction}

\begin{figure}[H]
		\centering
		\fbox{
			\begin{minipage}{0.8\textwidth}
				\begin{center}
					\textbf{Source text:}\\
					 The steel beam has a length of 11 inches and is oriented vertically.
					\vspace{-0.5em}
					\begin{center}
						\begin{tikzpicture}
							\draw[->, thick] (0,0) -- (0,-0.5);
						\end{tikzpicture}
					\end{center}
					\textbf{Extracted Attribute for the beam:} \\
					Material: Steel \\
					Length: 11 inches \\
					Orientation: Vertical
				\end{center}
			\end{minipage}
		}
		\caption{Example of attribute extraction.}
		\label{fig:attribute_value_extraction_example}
	\end{figure}
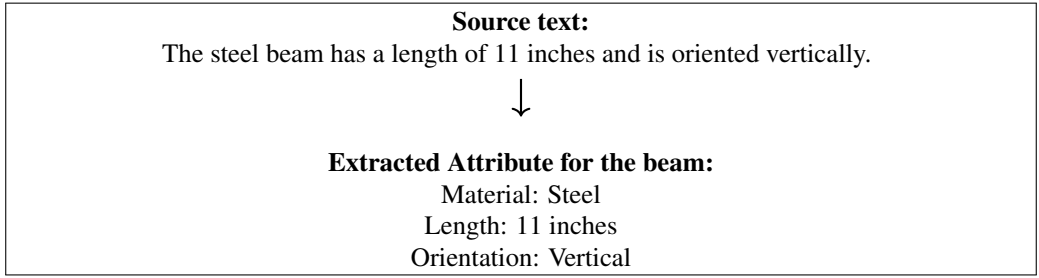

For each attribute identified in a component, quantitative (e.g., dimensional) or qualitative information (e.g., orientation) can be provided in the source text, this information is often referred to as its ``value'' (and it is not to be confused with a numerical value). When quantitative information is provided, attributes of each component can be thought as described by three parts that need to be extracted: (i) the attribute label; (ii) the attribute numerical value; (iii) the units of the attribute (if any). When qualitative information are provided, only two parts need to be extracted: (i) the attribute label; (ii) a descriptive phrase.  
For example, in \Cref{fig:attribute_value_extraction_example}, for the attribute label  ``length", the attribute value is ``11" and the unit of measurement is ``inches". For the attribute label ``orientation'' the descriptive phrase ``vertical'' needs to be extracted. 

	Although the distinction in quantitative and qualitative attributes is not critical in terms of the automated extraction strategy to be implemented, and it is in general referred to as ``attribute value extraction'',  this is critical at the validation stage of each BDD generated to ensure a sufficiently accurate representation of each component. 
    Typical approaches for attribute value extraction are broadly divided in those based on supervised machine learning methods, and those based on LLM  methods \cite{brinkmann2025extractgpt,zheng2018opentag}. The supervised learning methods need to be trained on a large corpus of labelled data, which is not available for the current task. Hence, LLM based methods were explored.  Specifically, the Llama 3.2 3B Instruct \cite{roziere2023code} model was used with one shot prompting to extract both quantitative and qualitative attributes. The Instruct model was used as it was tuned more appropriately for the attribute value extraction task. The one-shot prompt used is shown in \Cref{lst:attribute_value_extraction_prompt}.
    
	\subsubsection{Relationship Mapping}\label{sec:rel_mapping}

	Relationships in BDD diagrams can be categorised broadly into: composition, association, aggregation and generalization \cite{friedenthal14practical,omg19omg}. In order to identify the dynamical system component (i.e. a specific block) relationships with other components (i.e. other blocks) dependencies, composite association (i.e., a block is part of another block), generalization (i.e., one block inherites properties and behaviors from another block), and connection between blocks, the so-called reference association, are considered. The mapping converts these relationships extracted into lines and arrows in the BDD diagram. 
    To carry out the relationship mapping, the steps considered in \cite{zhong2023natural} are implemented:

	\begin{enumerate}
		\item \textbf{Identification of operation}: For each subject phrase $p_{\text{sub}}$ identified in textual relationships, the relation phrase represents the operation able to be performed by the component represented by the block defined by $p_{\text{sub}}$.
		\item \textbf{Classification based on relation phrase}: A list of pre-determined WordNet synsets (a group of words that are synonyms) that define composite relations (e.g. words like ``include" or ``composed of") is used to classify the relationship as a composite relationship. The synset used in this paper is ``include.v.01". If the relation phrase is in this list, the relationship is classified as a composite relationship, where the block defined by the object phrase is part of the block defined by the subject phrase.
		\item \textbf{Classification based on overlap}: The remaining textual relationships are classified by the overlaps between the string representations of the subject and object phrases. If the string representation of one of the phrases is contained in the other (e.g. ``rudder wire" and ``wire"), the shorter phrase should represent a more general version of the component represented by the longer phrase. Hence, the relationship is classified as a generalization relationship.
		\item \textbf{Classification based on importance metric}: The remaining textual relationships are classified according to the importance metric $\Gl_{p,d}$ defined in \Cref{eq:importance_metric}.If the difference between the importance metric of the subject and object phrases is greater than a user defined threshold $\Gs_{\text{rel-difference}}$, then the relationship is classified as a composite relationship, where the lower-importance phrase is the a part of the other phrase.
		\item \textbf{Classification of remaining relations}: The remaining relationships are classified as reference relationships, where the block defined by the subject phrase references the block defined by the object phrase.
	\end{enumerate}

	\subsubsection{Augmentation} \label{sec:Augmentation}

	At this stage, the generated BDD diagram is augmented to be more connected by exploiting the semantic relationships between the blocks. This process adds augmented relationships between the blocks and also additional augmented blocks to the BDD diagram.  Block augmentation follows the following steps:

	\begin{enumerate}
		\item \textbf{Identification of top-level phrases}: A sub-block is a block that is either at the part end of the composite relationship or a specialised end of a generalization relationship as described in \Cref{sec:rel_mapping}. Top level phrases are phrases whose blocks are not sub-blocks. Identify these top level phrases by looping through all the blocks using a depth-first search \cite{tarjan1972depth}.
		\item \textbf{Abstraction}: Multi-word top level phrases can be further generalised into shorter, and more general phrases, all the way till one word remains. This is done using \Cref{alg:phrase_abstraction} where the word with the lowest secondary importance metric $\gamma_{p,d}$ from the phrase is removed. The secondary importance metric $\gamma_{p,d}$ is defined as the sum of the tf-idf score and the normalized WordNet score of the words in the phrase. The notation for the tf-idf and the normalized Wordnet score has been used as described in \Cref{sec:key_phrase_select}. The $ p \setminus t $ operation removes the word $t$ from the phrase $p$. $p_{\text{abstract}}$ represents the abstracted phrase obtained by removing the word with the lowest $\gamma_{p,d}$ from the original phrase $p$.
		\item \textbf{Relationship Augmentation}: Now all of the top-level phrases are unigrams (``one-word phrases"). Augment the relationships between the blocks by adding relationships between the top-level phrases and the unigrams. This is done by using Wordnet to identify hypernym/hyponym and meronym/holonym relationships between the top-level phrases. The hypernym/hyponym relationship is used to identify generalization relationships (e.g. ``vehicle" as the hypernym and ``car" as the hyponym), while the holonym/meronym relationship is used to identify composite relationships (e.g. ``car" as the holonym and ``tyre" as the meronym).
		\item \textbf{Phrase Augmentation}: The lowest common hypernyms among the resultant top-level phrases is found using Wordnet. Blocks corresponding to these phrases are added to the diagram and generalization relationships between the newly formed blocks and blocks that have phrases with hyponyms to them are assigned. 
	\end{enumerate}

	\begin{algorithm}
	\caption{Phrase Abstraction step sequence adapted from \cite{zhong2023natural}}
	\begin{algorithmic}[1]
	\State Identify set of top-level phrases $H$ from document $d$ using corpus $c$
	\For{phrase $p \in H$}
		\If{length of phrase $N_p > 1$}
			\State Initialise score set $\Gamma$
			\For{noun $t \in p$}
				\State $\gamma_{t,d} = \text{tf-idf}_n(w, d, c) + h'_{t,d}$
				\State $\Gamma_{p,d} = \Gamma_{p,d} \cup \{\gamma_{t,d}\}$
			\EndFor
			\State $\text{sort}(\Gamma_{p,d})$
			\State $p_{\text{abstract}} = p \setminus t$, where $\gamma_{t,d}$ is smallest in $\Gamma_{p,d}$ \Comment{The $ p \setminus t $ operation removes the word $t$ from the phrase $p$.}
			\State $H = H \cup \{p_{\text{abstract}}\}$
		\EndIf
	\EndFor
	\end{algorithmic}
	\label{alg:phrase_abstraction}
	\end{algorithm}

\section{Steps for implementation and validation of context-free automated generation of SysML diagrams}\label{sec:ImplAndValidModels}

The steps for the automatic generation of the BDD diagrams to enable the automated evaluation of a dynamical system computational model can be summarized as follows:

\begin{enumerate}
\item Manually select the corpus of documents describing the class of dynamical systems that will be used to extract information and automatically build a computational model. Typically, this would consist of books, engineering standards, specifications, and/or technical documents. 
\item Manually select or write a document describing a specific configuration of the dynamical system to obtain the BDD diagram for the specific system of interest;

\item Specify the hyperparameter values for the process. A table with the hyperparameter values, their valid ranges, a description of their use and the values used in this paper is shown in \Cref{tab:recommended_hyperparameters}.
\item Select the phrase to define the parent block in the BDD diagram (optional). This step is used when the full BDD diagram is too large to be viewed directly and a sub-diagram parented by the block with the selected phrase will be shown instead;
\item Run the end-to-end approach as described in \Cref{sec:SysMLg} to automatically generate BDD diagrams, and that will yield the following intermediate outputs:
    \begin{enumerate}
    \item List of key nouns;
    \item List of extracted relationships;
    \item List of key phrases and key relationships;
    \item Attributes values extracted per block;
    \item BDD diagram generation by plotting the blocks as nodes and relationships as lines using graph visualization tools.
    \end{enumerate}
\item Validation of the quality of the resulting BDD for building a computational model through the following steps:
    \begin{enumerate}
    \item Evaluation of key phrase selection;
    \item Evaluation of the BDD diagram generated;
    \item Evaluation of the attribute values included in the BDD diagram.
    \end{enumerate}
     The user can use these metrics and scores in place of manual inspection of the source text.
\item  Based on metrics and scores obtained for partial outputs and on the resulting BDD diagram, the user can intervene to remove non-essential information. This would lead to a more refined BDD diagram.  
\end{enumerate}

The validation step is described in detail in what follows.

\subsection{Validation of the BDD diagrams generated}\label{sec:validationsys}
Two strategies are proposed to validate the results of the proposed BDD diagram generation strategy. The first focuses on the accuracy of the key phrase extraction as they form the blocks in the BDD diagram. Based on  observations on the information held by the BDD diagram generated, a validation strategy on the components of the BDD diagram is then proposed. This enables the quantification of similarity scores on  set of blocks, blocks, and attributes, together with match scores enabling the identification of the causes of poor similarities.  Results are validated against manually extracted results (considered as ground-truth), and compared against LLMs. Different case studies are investigated to highlight strengths and opportunities for improvement of the proposed strategy.

\subsubsection{Accuracy of key phrase extraction}

The ground truth for this validation comprises of manually extracted key phrases from Chapter 7 in \cite{jackman1912flying}, and results obtained with GPT-4o are also used as a benchmark. The performance metrics investigated are:
	\begin{align}
		\text{Precision} = \frac{\text{Exact Match}}{\text{Total Extracted Phrases}} \label{eq:precision} \\
		\text{Recall} = \frac{\text{Exact Match}}{\text{Total Ground Truth Phrases}} \label{eq:recall} \\
		\text{Fuzzy Precision} = \frac{\text{Fuzzy Match}}{\text{Total Extracted Phrases}} \label{eq:fuzzy_precision} \\
		\text{Fuzzy Recall} = \frac{\text{Fuzzy Match}}{\text{Total Ground Truth Phrases}} \label{eq:fuzzy_recall}
	\end{align}
    
		The term ``exact match" refers to the number of key phrases that were extracted that exactly match a ground truth key phrase, while ``fuzzy match" refers to the number of key phrases that were extracted that have any overlap with a ground truth key phrase. An overlap is when any word in the ground truth key phrase is present in the extracted key phrase. For example, ``rudder beam" (ground truth key phrase) overlaps with ``frame beam" (extracted key phrase) as both phrases have the word ``beam".  The term ``normalized extracted phrase number" refers to the total number of extracted key phrases divided by the total number of ground truth key phrases. This value is always greater than 0. When this number is greater than one, the algorithm has extracted more key phrases than the ground truth. It is beneficial to have a value greater than one for this number as non-essential phrases can be removed by the user. The term ``precision" refers to the fraction of extracted key phrases that are in the ground truth, while ``recall" refers to the fraction of ground truth key phrases that are extracted.

\begin{figure}
		\centering
		\includegraphics[width = 0.8\textwidth]{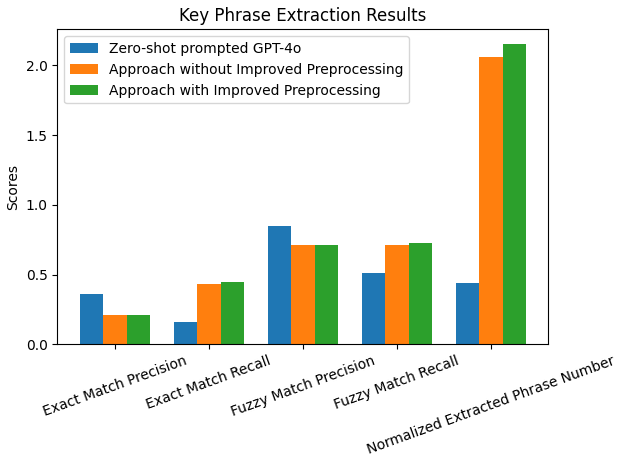}
\caption{Comparison of key phrase extraction performance of:  zero-shot prompted GPT-4o,  proposed approach with preprocessing improvements and approach without preprocessing improvements \cite{zhong2023natural}}
		\label{fig:preprocessing_improvements}
	\end{figure}

The validation results of the key phrase extraction are shown in \Cref{fig:preprocessing_improvements}. The results show that the current methodology outperforms GPT-4o\footnote{The prompts for this can be found at https://chatgpt.com/share/67771274-b768-800f-9ef4-755172c091de} in key phrase extraction on recall (fraction of ground truth phrases extracted) whilst performing slightly worse on precision (fraction of extracted phrases that are in the ground truth). The results also show that the preprocessing improvements marginally improve the key phrase extraction performance, more notable in terms of recall.

    \subsubsection{Observations on the information encoded in the BDD diagrams} \label{sec:observe_BDD}

The information encoded in the  automated BDD diagram generated can be directly compared with the source text that was used to generate the BDD. Alternatively, the automated BDD diagram can be compared with the manually-generated BDD diagram, and/or with the zero-shot prompted GPT-4o generated BDD diagram, to draw conclusions on the information encoded in  the BDD diagram generated, and to investigate which information should be used to assess the accuracy of this information. Two type of systems were investigated, for which two different corpus of documents were considered. 
Since BDD diagrams, even of simple systems, can be very complex, instead of comparing complete BDD diagrams, the sub-diagrams formed by taking a specified block and all the connected sub-blocks under it were compared. The observations made in this subsection form the basis for the scores  proposed in \Cref{sec:attr_val_extract}. 

The first example is focused on a patent document on the ventilation of internal combustion engines \cite{yasuda2021jp6875871b2}, where information are expressed in context-rich images and in text. This is particularly challenging for the proposed approach since only text is retained and images are discarded. The patent document \cite{yasuda2021jp6875871b2} whose excerpt can be found in \Cref{tab:document_types} was used, together with a corpus of 10 patents. The generated BDD sub-diagram under the ``turbine" block is shown in \Cref{fig:patent_bdd}. Here the information extracted from the generated BDD diagram is directly compared with the source text (The comparison is done on extracts where the text corresponds to the components in the sub-diagram as the sub-diagram is not a complete representation of the entire system described by the patent).  It can be observed that the BDD diagram successfully extracts meaningful components of the system as blocks, as components such as ``turbine" and ``turbine housing" which are components mentioned in the source text were extracted.
However, due to the many numbers in the text indicating the components in the figures of the patent document (e.g. ``The exhaust valve cam 46a is in contact with the end of the exhaust valve 40,"), the BDD diagram incorrectly identifies some of them as attributes. This is due to the inability of the current methodology to extract information from the context-rich images in the patent. The numerical values in the source text are, therefore, incorrectly identified as attribute values for the components of the system In the context of engineering practices, it might be time consuming and error prone, to check the information encoded in the entire BDD diagram generated by going through the entire source document. This example supports the use of multimodal information extraction for generating the BDD diagrams, which is beyond the scope of the present work. 

  Alternatively, a direct and relatively straightforward comparison can be made on the extracted blocks and relationships between the extracted blocks for a specific sub-diagram of interest to the engineer, and this could potentially be automated. When the focus is the one-to-one block comparison, attention must be given to the semantic similarity of the labels of the blocks extracted and all the parts and operations sections of the blocks.
  
For example, let us consider the manually generated BDD sub-diagram from Chapter 7 in \cite{jackman1912flying} under the parent block ``glider" is shown in \Cref{fig:ground_truth_bdd}  with the rest of the book as the corpus. It is possible to observe 8 blocks with composite and generalization relationships between them.     
The BDD diagram extracted by using the proposed approach is shown in \Cref{fig:provided_bdd}. By comparing the two diagrams, it can be seen that the rudder, rudder beam and glider blocks are present in the automatically generated BDD diagram as is present in the ground truth. The front and rear rudders were not extracted as blocks but were extracted as attributes under the rudder block. The remaining blocks in the ground truth were not present in the generated BDD diagram.   The ``manipulated" operation of the rudder was successfully captured by the generated BDD diagram. None of the parts in the parts section were accurately captured by the automatically generated BDD diagram. None of the relationships were accurately captured either as very little of the blocks were generated successfully.
    
Similarly, the BDD diagram generated with GPT-4o was zero-shot prompted \footnote{The prompts for this can be found at https://chatgpt.com/share/6741eb41-ccf4-800f-86d6-12cc6ae1c334} for the same text, is shown in  \Cref{fig:gpt4_bdd}. 	Overall, the GPT results took more prompting to generate diagrams. The GPT generated diagram successfully extracted the rudder and rudder beam block, whereas all the other blocks were not successfully extracted. This is slightly less than the successfully extracted blocks in the proposed approach, with the difference being the extraction of the ``glider" block. The GPT generated diagram included a variety of different relationships extracted not following the predefined list of generalization, composite, and reference relationships that we have restricted ourselves to in our approach (e.g. ``Reinforced By"). This would make the conversion of the BDD diagram into code deterministically using methods described later in \Cref{sec:skeleton_gen} much harder, as deterministic mappings from relationship type to code was done. 

In what follows, similarity scores and match scores and introduced to automatically quantify the accuracy of blocks and attributes of a generated BDD diagrams, enabling a systematic evaluation of BDD diagrams.

	\subsubsection{Accuracy of blocks and attributes from the extracted BDD Diagram} \label{sec:attr_val_extract}

    The attribute value extraction is typically assessed with the precision, recall and F1 score by directly comparing attributes extracted from the same sentence \cite{xu2019scaling}.
    To the best of the authors' knowledge, the assessment of the accuracy of the blocks and attribute value extraction of multiple components in dynamical systems has not been addressed in the literature.
    
It is proposed to automatically compare the extracted blocks and attribute values of all the components in the entire output BDD diagram (for example the one in \Cref{fig:bdd_attr_grid})  with the manually generated BDD diagram (for example, the one in \Cref{fig:bdd_attr_grid_ground_truth})  and to provide a score to quantify the accuracy of the attributes extracted. It is worth noting that the relationships (arrows) between blocks were omitted to simplify the BDD  diagrams, since the relationships are not part of this validation stage. 

From the previous sections it can be concluded that validating blocks and attributes extraction requires assessing: (i) block numbers: the automatically generated BDD diagrams might have less or more blocks compare to the manually generated one (e.g., because of inaccurate coreference resolution); (ii) block labels: inaccurate/imprecise labels assigned to a block referring to a specific component of the dynamical system; (iii) number of attribute labels: inaccurate/imprecise number of attribute values identified within the same block; (iv) attribute label: inaccurate/imprecise label assigned to an attribute; (iv) attribute value or description: inaccurate/imprecise/missing attribute numerical value or descriptive value; (v) attribute unit: inaccurate/imprecise/missing  assigned unit of the attribute. 
    When comparing numerical values (i.e., (i) and (iv)), the numerical values are converted into stringified labels which are then mapped onto vectors so that the same strategy as that applied to labels can be used.  When comparing labels, either at the block or at the attribute level, the labels can be mapped into vector descriptions and the cosine similarity can be used to quantify the similarity between the two vectors \cite{jurafsky2009speech}.  Two labels are close to each other in the vector space if they have the same semantic meaning.  The cosine similarity $CS$ is defined as:






	\begin{align}
		CS(\textbf{a},\textbf{b}) = \frac{\textbf{a} \cdot \textbf{b}}{||\textbf{a}|| \, ||\textbf{b}||} \label{eq:cosine_similarity}
	\end{align}
    
	where $\cdot$ is the dot product between two vectors. The $CS$ ranges from -1 to 1, where two identical labels will produce a result of 1 (the two vectors point in exactly the same direction) and two completely different labels will produce a result of -1 (the two vectors point in opposite directions).  Three quantities for defining similarities scores are introduced in what follows. These will be used to specify similarity metrics. We also refer to semantic similarity as how closely related the semantic meaning is between two strings, which is quantified by the $\text{CS}$ of their mapped vectors. For this section, when we notate strings $a$ with boldtext $\textbf{a}$, the bolded versions represent the vectors that the strings are mapped to.

\textbf{Attribute Similarity score:}
The attribute similarity is defined as the semantic similarity of an attribute's values and units with another attribute's values and units. This attribute similarity should only be compared between two attributes whose attribute labels are sufficiently semantically similar for a valid comparison.
 Let us indicate the attributes being compared as $a_1$ and $a_2$, the attribute values being compared as $v_1$ and $v_2$, and similarly, the attribute units being compared with $u_1$ and $u_2$. 
The attribute similarity between the two attributes includes information from the label and the value and is defined as:

	\begin{align}
	\text{attribute similarity}(a_1,a_2) = \begin{cases}
		CS(\textbf{v}_1,\textbf{v}_2) & \text{if } u_1 = u_2 \\
		0.5 \times (CS(\textbf{v}_1,\textbf{v}_2) + CS(\textbf{u}_1,\textbf{u}_2))& \text{if } u_1 \neq u_2 \\
	\end{cases} \label{eq:attribute_similarity}
	\end{align}

The attribute similarity score ranges between 0 and 1. The attribute labels are not considered in the evaluation of this score as in this algorithm, attributes will be matched by the semantic similarity of their attribute labels before a comparison is being made as will be described in the following text. This essentially guarantees that attributes with semantically similar labels are compared throughout the algorithm.

\textbf{Block Similarity score:}
	The block similarity is defined as the average of the attribute similarities between the attributes in the block and the attributes in the ground truth block. The attribute match function is used to find the best matching attribute in the ground truth block for each attribute in the block.  
    Let $B$ denotes a block which has a set of attributes, and similarly, $G$ denotes a ground truth block which has a set of attributes.  The similarity between the two blocks which ranges between 0 and 1 is defined as:
	\begin{align}
	\text{block similarity}(B,G) = \frac{1}{|G|} \sum_{a_g \in G} \text{attribute similarity}(a_g,a_{\text{match}}(a_g, B) ) I(a_{\text{match}}\neq \text{None})\label{eq:block_similarity} \\
	a_{\text{match}}(a_g, B) = \begin{cases}
		\displaystyle {\argmax_{a_b \in B}}  \ CS(label(a_g), label(a_b)) & \text{if } CS(a_b,a_g) > \Gs_{\text{attribute}} \\
		\text{None} & \text{otherwise} \\
	\end{cases} \label{eq:attribute_match}
	\end{align}

	where  $| \cdot |$  is an operator that for each block $G$ would return the number of attributes within that block and the $label$ function extracts the label of the given attribute or block (depending on input) and outputs its vector mapping. The symbol $a_g$ represent one attribute in the  ground truth block $G$, $a_{\text{match}}$ is the corresponding attribute in the corresponding extracted block $B$, $I(\cdot)$ is an indicator function and $\Gs_{\text{attribute}}$ is a thresholding value (that is set as 0.7). 

\textbf{Set Similarity score:}
	Mathematically, a set is a collection of elements \cite{hausdorff2021set}, and for this section we will compare sets of blocks from the BDD diagram. The set similarity is defined as the average of the block similarities between the blocks in the set and the blocks in the ground truth set that have non-zero attributes. This is normalized by the number of  blocks in the ground truth set that have non-zero attributes. The set similarity behaves similarly to a recall metric for the sets, due to the normalization of the ground truth set's non-zero attribute blocks. It is worth noting that having two zero-attribute blocks match would automatically give a maximum block similarity of 1. This would inflate the score of the set similarity without any attributes actually being compared. Consequently, the blocks with zero attributes were not included in the comparison, as specified in the definition provided.   The accuracy by which the zero-attribute blocks are extracted is defined in a separate score explained later in this section.
    
    The block match function is used to find the best matching block in the ground truth set for each block in the set. Let $\mfB$ denote a set of blocks and $\mfG$ denote a ground truth set of blocks. The similarity between the two sets of blocks that ranges between 0 and 1 is defined as:

	\begin{align}
	\text{set similarity}(\mfB,\mfG) &= \frac{1}{\sum_{g \in \mfG} I(|g|>0)} \sum_{g \in \mfG} \text{block similarity}(g,b_{\text{match}}(g, \mfB))I(|b_{\text{match}}(b, \mfG)| \neq \text{None} \cap |g| > 0) \label{eq:set_similarity} \\
	b_{\text{match}}(g, \mfB) &= \begin{cases}
		\displaystyle \argmax_{b \in \mfB} \ CS(label(b),label(g)) & \text{if } CS(b,g) > \Gs_{\text{block}} \\
		\text{None} & \text{otherwise} \\
	\end{cases} \label{eq:block_match}
	\end{align}

	where $b$ is a block in the set $\mfB$, $g$ is a block in the ground truth set $\mfG$, $\Gs_{\text{block}}$ is a thresholding value (that is set as 0.7). 

\textbf{Match scores:} Two match scores are introduced to identify other sources of poor similarity not captured by the set similarity scores. One score is 
the number of zero-attribute ground truth blocks correctly extracted as zero-attribute blocks. This value is normalized by the number of zero-attribute ground truth blocks that have a corresponding extracted block. The normalization restricts the range of the score between 0 and 1.
 This is called the ``zero attribute match score''. This score is important as it ensures that when the extracted blocks don't unnecessarily extract attributes when they should not exist for the system component that they are representing.

The zero attribute match score is defined as:
	\begin{align}
	\text{zero attribute match score}(\mfB,\mfG) = \begin{cases}
    1 & \sum_{g \in \mfG }I(|g| = 0) = 0 \\    
	\frac{\sum_{g \in \mfG} I( |g| = 0 \cap b_{\text{match}}(g, \mfB) \neq \text{None} \cap |b_{\text{match}}(g, \mfB)| = 0)}{\sum_{g \in \mfG }I(|g| = 0) \cap b_{\text{match}}(g, \mfB) \neq \text{None})} & \text{otherwise}
    \end{cases}
    \label{eq:false_positive} 
	\end{align}

	where $b_{\text{match}}(g, \mfB)$ is the best matching block in the extracted set for the ground truth block $g$. When the score is 0, all of the ground truth blocks that have no attributes matched with extracted blocks with attributes. When the score is 1, all the of the ground truth blocks that matched with blocks with zero attributes in the extracted set don't have attributes themselves too.

The ``normalized set match score''is introduced to quantify how well the extracted block labels match with the ground truth block labels. The normalized set match score is defined as:

    \begin{align}
        \text{normalized set match score}(\mfB,\mfG) =  \frac{1}{|\mfG|} \sum_{g \in \mfG} I(b_{\text{match}}(g, \mfB) \neq \text{None})
    \end{align}

This score can take values ranging from 0 to 1. When the score is 0, this means that none of the ground truth block labels have a match with the any extracted set block labels. When the score is 1, all of the ground truth block labels match with the extracted set block labels.

\textbf{Application of the proposed similarity scores and match scores:} Blocks and attributes extracted from Chapter 7 in \cite{jackman1912flying}, are shown in \Cref{fig:bdd_attr_grid} - where since the focus is attribute value extraction, and not relationship, the lines and arrows are omitted. The corresponding ground-truth BDD diagram manually obtained is shown in \Cref{fig:bdd_attr_grid_ground_truth} . The results obtained with the proposed similarity and match scores are shown in \Cref{tab:flying_machine_normalized_metrics}.

\begin{table}[h!]
	\centering
	\begin{tabular}{|l|c|}
		\hline
		\textbf{Similarity/match score} & \textbf{Score} \\  \hline
       Block Similarity: rudder - rudder & 0.00 \\ \hline
       Block Similarity: rudder beam - rudder beam & 0.85 \\ \hline
       Block Similarity: cross piece - cross piece & 0.00 \\ \hline
		Set Similarity & 0.21 \\ \hline
		Zero Attribute Match Score & 0.75 \\ \hline
        Normalized Set Match Score & 0.36 \\ \hline
	\end{tabular}
	\caption{Results for BDD diagram auotmatically extracted (\Cref{fig:bdd_attr_grid}) versus the BDD ground truth (\Cref{fig:bdd_attr_grid_ground_truth}).}
	\label{tab:flying_machine_normalized_metrics}
\end{table}   

The results show that the block similarity of the rudder beam and its matched block is very high, indicating that the attributes between the blocks are very similar. The remaining blocks however did not have similar attributes to one another. However, overall the attribute value extraction is not very accurate as only one block had its attributes matched well with its corresponding ground truth block. The zero attribute match score was high, meaning that the algorithm did not unnecessarily extract attributes for blocks that did not have attributes. The normalized set match score was quite low, meaning that the extracted block labels did not match well with the ground truth block labels.

The low rate of correctly identified attributes and bad matching of block labels could be due to the older language in the document. Hence, a more modern sample text describing a hydraulic robotic arm system was generated using GPT-4o and investigated the performance of the attribute value extraction on this text. And excerpt of the sample text is shown in \Cref{fig:hydraulic_text}. The results of the attribute value extraction on this text are shown in \Cref{tab:normalized_metrics}. The results show that the attribute value extraction is slightly more accurate than the previous text, marginally confirming our hypothesis that it is easier to extract attributes of components from modern text. The zero attribute match score for this text is 1 mainly because all the ground truth blocks have attributes in them. The normalized set match score was much higher than the previous text, showing that the extracted block labels match very well with the ground truth block labels. Overall, the scores indicate an improvement with respect to the algorithm without this step as at least some attributes of the system components are extracted.

\begin{figure} [h!]
	\centering
		\fbox{
			\begin{minipage}{0.8\textwidth}
				\begin{center}
					The Hydraulic Robotic Arm System (HRAS) is designed for precision handling and heavy-duty industrial applications. It consists of several key components that enable smooth operation, high load capacity, and real-time control.The base frame is constructed from high-strength aluminum alloy, providing a structural foundation with a load capacity of 500 kg. It is mounted on a 360-degree rotating platform with a rotational speed of 30 degrees per second, allowing full-range motion. $\cdots$
				\end{center}
			\end{minipage}
		}
		\caption{Excerpt of synthetic text on a hydraulic robotic arm system.}
		\label{fig:hydraulic_text}
\end{figure}

\begin{table}
	\centering
	\begin{tabular}{|l|c|}
		\hline
		\textbf{Similarity/match score} & \textbf{Score} \\ \hline
        Block Similarity: base frame - base frame & 0.58 \\ \hline
		Block Similarity: actuator system - actuator system & 0.93 \\ \hline
		Block Similarity: cylinder - cylinder & 0.50 \\ \hline
        Block Similarity: sensor network - sensor network & 0.00 \\ \hline
        Block Similarity: control module - control module & 0.00 \\ \hline
        Block Similarity: power system - power system & 0.00 \\ \hline
        Block Similarity: hydraulic pump - foundation load & 0.00 \\ \hline
		Set Similarity & 0.29 \\ \hline
		Zero Attribute Match Score & 1.00 \\ \hline
        Normalized Set Match Score & 0.86 \\ \hline
	\end{tabular}
	\caption{Results for Synthetic text with preprocessing improvements}
	\label{tab:normalized_metrics}
\end{table}

The performance of the attribute value extraction on a synthetic text describing a hydraulic system was also assessed without the improved preprocessing steps. The results are shown in \Cref{tab:normalized_metrics_hydraulic}. The results show that the attribute value extraction is marginally less accurate than the previous example. However, the zero attribute match score and normalized set match score stayed the same. This demonstrates the importance of improved preprocessing steps in the attribute value extraction process.

\begin{table}
	\centering
	\begin{tabular}{|l|c|}
		\hline
		\textbf{Similarity/match score} & \textbf{Score} \\ \hline
        Block Similarity: base frame - base frame & 0.58 \\ \hline
        Block Similarity: actuator system - actuator system & 0.93 \\ \hline
		Block Similarity: cylinder - cylinder & 0.50 \\ \hline
		Block Similarity: sensor network - sensor network & 0.25 \\ \hline
        Block Similarity: control module - control module & 0.00 \\ \hline
        Block Similarity: power system - power system & 0.00 \\ \hline
        Block Similarity: hydraulic pump - foundation load & 0.00 \\ \hline
		Set Similarity & 0.19 \\ \hline
		Zero Attribute Match Score & 1.00 \\ \hline
        Normalized Set Match Score & 0.86 \\ \hline
	\end{tabular}
	\caption{Results for synthetic text without improved preprocessing steps}
	\label{tab:normalized_metrics_hydraulic}
\end{table}

\section{From SysML diagrams to dynamical system computational models}\label{sec:Sysmltomodel}

The task of creating computational models from SysML diagrams is typically carried out manually by using XMI (XML Metadata Interchange) to add relevant metadata (such as equations) from a database of relevant information into the SysML to form a meta-model. The meta-model is then parsed to generate a script in the target simulation language \cite{huang2007system}. To the best of the authors' knowledge, it is currently not possible to  automatically generate the meta-model of a general dynamical system from the SysML diagram.

As schematically depicted in \Cref{fig:overview_flowchart}, it is proposed to carry out two steps for generating large scale computational models of complex dynamical systems from SysML diagrams: code generation, and computational model generation. As for the automate SysML generation step, the user can directly intervene on the results of each of these step to improve the overall model performance.  These steps and their outputs are detailed in what follows.

\subsection{Code Generation}

 The code generation stage will return as output class files modeling each component in the BDD diagram according to the underlying dynamical system, organised in a hierarchical file structure. All the files will be written in a chosen programming language. The language chosen for the work presented in this paper is Python, for simplicity.
 The code generation stage is comprised of file structure generation, followed by file structure generation, class skeleton generation and code comment generation and finally function body generation. 

 The code generation process is made automatic by the use of a templating engine. A template is a blueprint for autogenerated code where parts of the code can be set as parameters to be passed into the template to complete the code. The process of passing parameters into a template and automatically generating the code is called rendering. The templating engine used for this project is Jinja2 \cite{ronacher2008jinja2}.

	\subsubsection{File Structure Generation}

 This step concerns the generation of a file structure that helps organize the generated code for each component the computational model. A good file structure is one that mimics the  structure of the dynamical system itself. The dependencies of components can be extracted from the composition relationships in the BDD diagram. Specifically, a block that is a part of a parent block should have its code generated in a file living in a subdirectory of the file representing the parent block that it is a part of. This approach generates a folder hierarchy with blank files inside them to be populated in further steps. 	The following is the sequence of steps are automated to generate the file structure from a BDD:
	\begin{enumerate}
		\item Start with a set of blocks from the BDD diagram. To ensure each block is generated as a single unique file, Remove blocks from the set whenever the file for the corresponding block has already been generated. 
		\item Identify the top level blocks in the set. These blocks are the ones that  are not children of any other composite relationships.
		\item For each top level block, automatically create a blank file that will be populated with a code corresponding to the function of that block. Code file names are the same as their block labels. If the block is an augmented block, ignore the creation of its file. This is because augmented blocks are not based on sentences in the actual text and is artificially generated in the augmentation stage (described in \Cref{sec:Augmentation}) based on the semantic meanings of the block labels, they only serve as tools to make the BDD diagram more connected visually but serve no purpose in the computational modeling of the dynamical system.
		\item Create a subdirectory for all the top level blocks that have composite children blocks. The subdirectory name is the label of the top level block with the suffix ``\_parts".
		\item Within each subdirectory, create a file for each child block and other blocks that have non-composition relationships with them.
		\item Repeat this process by treating the blocks in the subdirectory as top level blocks and creating subdirectories for them. This is done recursively until the set of bloks is empty.
	\end{enumerate}

	An example of the file structure generated from the BDD diagram in \Cref{fig:simple_pendulum_bdd} is shown in \Cref{fig:pendulum_folder_structure}. 

    \subsubsection{Class Skeleton Generation} \label{sec:skeleton_gen}

	For each file generated, the code is organised according to the object oriented programming paradigm \cite{rentsch1982object}. The class skeleton for each file is automatically generated by:  using information within each block of the BDD diagram to define  class label, class variables and operations; using the relationships between the blocks as class relationships. Specifically, the class skeleton is automatically generated by:

	\begin{enumerate}
		\item For each block in the BDD diagram, a class with the same name as the block label is automatically generated. The class will contain the attributes of the block as class variables. This class will live in the file bearing the same name as the block label.
		\item Depending on the relationships between blocks, the following class relationships can be automatically defined
            \begin{enumerate}
                \item  If the block is a generalization sub-block from another block, the class will inherit from the parent block class.
                \item If the block owns another block as part of its composite relationship, the class will instantiate the child block as a class variable.
            \end{enumerate}
		
		\item The initialization function (i.e. a function that creates an instance of the class) is the function called first during the construction of each instance of an object belonging to the class. The initialization function of the class, is populated by setting the values of the attributes of the block to the values extracted from the BDD diagram.
		\item For each operation detected in the block, a function with the same name as the operation is automatically created. The body for the function modeling the equation is not generated here and will be generated during function body generation which is described in \Cref{sec:comp_model_code_generation}.
		\item Thee import statements required for the class to work are automatically added at the the top of the file. This was done automatically by  determining deterministically which classes need to be imported by inspecting the generalization and composite relationships tied to the block being considered. For each generalization relationship, the file representing the block at the specific end of the relationship should import its general parent block's file from its file living on the same folder level. For each composite relationship, the file representing the block on the composite end of the relationship should import the child block's file living in a subfolder level relative to the current file. More imports can be included, as described in \Cref{sec:comp_model_code_generation}.
		\item Code comment generation: Docstrings for the class and the functions are automatically created and added as described in \Cref{sec:code_prompt_generation}. 
	\end{enumerate}

	An example of the code structure generated from the Motion block from the BDD diagram in \Cref{fig:simple_pendulum_bdd} is shown in \Cref{lst:Motion_class_example}.

	\subsubsection{Code Comment Generation}
	\label{sec:code_prompt_generation}

	To assist the user in understanding the code that is generated for a class, code comments are automatically generated for each class and function. These code comments are in the form of class and function docstrings that describe each class and function  (note again that the function body generation step is detailed in \Cref{sec:comp_model_code_generation}). The code comment generation can be broken down into two parts:

    \begin{itemize}
        \item \textbf{Extracting the context:} The context related to each class and each function needs to be  extracted first. For each class, the context is  the concatenation of all the sentences used to extract the class (i.e. the sentences that are used to extract the class name and the sentences that are used to extract the class attributes). 

    For example,  the sentence ``The rudder beam is 8 feet 11 inches long" is used to extract the attribute value ``length: 8 feet 11 inches" in the rudder beam block from \Cref{fig:provided_bdd} (sub-diagram of BDD diagram generated from Chapter 7 of \cite{jackman1912flying} under the ``glider" block). Therefore, this sentence will be included in the context of the class as input to be summarized. Similarly, the context of each function is  the sentence used to extract the operation of BDD blocks corresponding to the function.
    
    In both cases, the user can specify which of the surrounding sentences should be included to add to the context by specifying a mask $\mathbf{m} = (m_{\text{before}}, m_{\text{after}})$, where $m_{\text{before}}$ represents the number of sentences before the sentence used to extract the operation to add to the context and $m_{\text{after}}$ represents the number of sentence after the sentence used to extract the opration to add to the context. For example, since the sentence `` The real motor-propelled flying machine, generally has both front and rear rudders manipulated by wire cables at the will of the operator." is used to extract the ``manipulated()" function of the rudder block from \Cref{fig:provided_bdd} (sub-diagram of BDD diagram generated from Chapter 7 of \cite{jackman1912flying} under the ``glider" block), therefore the sentence will be included as the context of the function as input to be summarized. If the user specified a mask of $\mathbf{m} = (1,1)$, then the sentence before and after this is also included in the context to be summarized.

	\item \textbf{Summarization tools:}
		The context extracted is used as an input into a summarization tool to generate the docstring for the class or function.         There are 2 main types of summarization tools that are typically used, extractive and abstractive summarization \cite{mahajani2019comprehensive}. Extractive summarization is the process of selecting a subset of sentences from the input text to form a summary. Abstractive summarization is the process of generating new sentences that capture the main ideas of the input text. The use of both types of summarization tools was investigated and the results showed that extractive summarization is more suitable for our task as it preserves the original meaning of the sentences and is less prone to hallucination.

	The extractive summarization tool used is the TextRank algorithm \cite{mihalcea2004textrank}. The TextRank algorithm is a graph-based algorithm that ranks the sentences in the input text based on their importance. The algorithm works by constructing a graph where each sentence is a node and the edges between the nodes represent the similarity between the sentences. The algorithm then uses a random walk approach to rank the sentences based on their importance. 	The example results of using Textrank on the sentences that were used to extract the class and function are shown in \Cref{fig:textrank_example}. 

	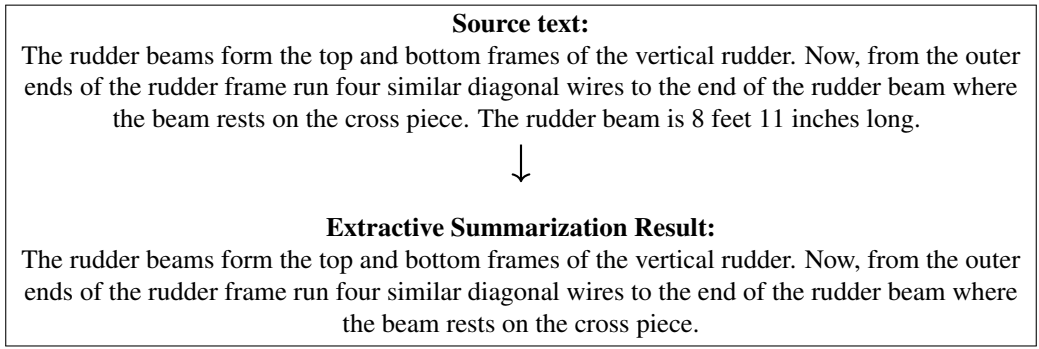
\begin{figure}
		\centering
		\fbox{
			\begin{minipage}{0.8\textwidth}
				\begin{center}
					\textbf{Source text:}\\
					 The rudder beams form the top and bottom frames of the vertical rudder. Now, from the outer ends of the rudder frame run four similar diagonal wires to the end of the rudder beam where the beam rests on the cross piece. The rudder beam is 8 feet 11 inches long.
					\vspace{-0.5em}
					\begin{center}
						\begin{tikzpicture}
							\draw[->, thick] (0,0) -- (0,-0.5);
						\end{tikzpicture}
					\end{center}
					\textbf{Extractive Summarization Result:} \\
					The rudder beams form the top and bottom frames of the vertical rudder. Now, from the outer ends of the rudder frame run four similar diagonal wires to the end of the rudder beam where the beam rests on the cross piece. 
				\end{center}
			\end{minipage}
		}
		\caption{Example of extractive summarization using TextRank on the docstring describing a rudder beam in a glider from text in Chapter 7 of \cite{jackman1912flying} .}
		\label{fig:textrank_example}
	\end{figure}

	The Pegasus extraction model \cite{phang2022investigating} was used to generate the abstractive summaries with different minimum lengths specified. The Pegasus model is a transformer-based model that is trained on a large corpus of text to generate abstractive summaries. The model works by generating a sequence of words that captures the main ideas of the input text. 	The results of using the Pegasus model to generate the abstractive summaries are shown in \Cref{fig:pegasus_example}. It is shown that, as anticipated, the model will hallucinate information if given the task to produce a summary longer than the original text (up to 60 words). However, even when the minimum length is set to 25, the model hallucinates information that is not present in the input text (there was no mention about the importance of the rudder in the input text), and that is not relevant for the modeler. This is a common problem with abstractive summarization models and is one of the reasons why it is better to use extractive summarization for our task. This problem only worsens when the minimum length is increased to 60, where the model generates a summary that is not even close to the input text.

	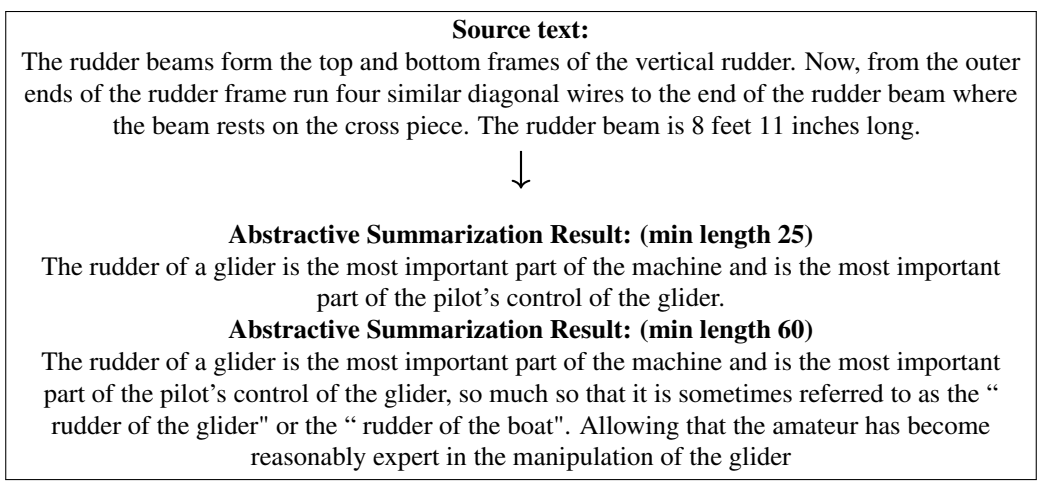
\begin{figure}
		\centering
		\fbox{
			\begin{minipage}{0.8\textwidth}
				\begin{center}
					\textbf{Source text:}\\
					 The rudder beams form the top and bottom frames of the vertical rudder. Now, from the outer ends of the rudder frame run four similar diagonal wires to the end of the rudder beam where the beam rests on the cross piece. The rudder beam is 8 feet 11 inches long.
					\vspace{-0.5em}
					\begin{center}
						\begin{tikzpicture}
							\draw[->, thick] (0,0) -- (0,-0.5);
						\end{tikzpicture}
					\end{center}
					\textbf{Abstractive Summarization Result: (min length 25)} \\
					The rudder of a glider is the most important part of the machine and is the most important part of the pilot's control of the glider.\\
					\textbf{Abstractive Summarization Result: (min length 60)} \\
					The rudder of a glider is the most important part of the machine and is the most important part of the pilot's control of the glider, so much so that it is sometimes referred to as the `` rudder of the glider" or the `` rudder of the boat". Allowing that the amateur has become reasonably expert in the manipulation of the glider
				\end{center}
			\end{minipage}
		}
		\caption{Example of abstractive summarization using Pegasus on the docstring describing a rudder beam in a glider from text in Chapter 7 of \cite{jackman1912flying}  .}
		\label{fig:pegasus_example}
	\end{figure}

	    \end{itemize}

	\subsection{Function Body Code Generation}
	\label{sec:comp_model_code_generation}

	After generating a structure for the computational model, the functions bodies in the classes are generated. A full treatment of this subject is out of the scope of the paper, and hence this paper provides a naive implementation that works on very simple systems to demonstrates the potential of the end-to-end process of the proposed methodology. The results for the methods discussed in this section are shown in \Cref{sec:validation}.

    It is worth mentioning that alternative LLM-based summarization tools could have been employed, but those are beyond the scope of the present work. 
    
	\subsubsection{Function Body Generation strategy} \label{sec:genBodyStrat}

	\label{sec:code_generation_methodology}
	The function implementation proposed is inspired by the paper  \cite{jiang2023multilingual} in which a method of inferring mathematical proofs from natural language text was proposed. Specifically, in \cite{jiang2023multilingual}, a database of informalized mathematical proofs (an informal description of mathematical proofs and their corresponding proofs) was created, and for the identification of a mathematical proof from natural language,  the similarity of the natural language prompt to the database to choose the mathematical proofs was used \cite{jiang2023multilingual}. 
    
    To perform function implementation, it is necessary to provide the following:

	\begin{enumerate}
		\item A set of (dynamical systems) equation implementation templates to choose from, with an informal description of the system, the states of the equations, and the necessary parameters for the equation. An example of the information required is show in \Cref{tab:dynamical_function_dict}, with an example implementation template written in Jinja2 (a templating language) shown in \Cref{lst:example_dynamical_equation_template}. This set of templates has to be manually compiled by engineers with domain knowledge on the system of interest.
		\item There are some model parameters that are  used for applications within the domain of the dynamical system  which are typically set to constant values  (e.g. if gravitational force is a commonly used term, it should be added to the database of constants). The labels values and values of such constants would be included in a database. This database would be manually prepared, but could be potentially automatically extracted from books. 
		\item A database showing the numerical attributes of other components in the system and the component holding the attribute. This is automatically extracted from the BDD diagram by looping through all of the blocks representing the components and extracting their numerical attributes.
	\end{enumerate}

	The function implementation is then automatically extracted by carrying out the following steps:

	\begin{enumerate}
		\item Compute the cosine similarity (see \Cref{sec:attr_val_extract}) between the code comment for the function whose function body is being generated and the list of informal descriptions of all the dynamical equations. Select the most similar description if its cosine similarity is above a tuned threshold  $\Gs_{\text{func}}$. The equation corresponding to the selected description will be implemented by filling in its implementation template.
		\item Automatically identify the parameters for the dynamical equation by using again the cosine similarity: (i) finding an exact match between the label of the parameter in the constant database, (ii)  using  the labels of the attributes of the current class; (iii) using the labels of the numerical attributes in the other classes. If a match is found, the value of the parameter is assigned to the matching value/attribute, otherwise the value of the parameter is set to a default value. The default value is usually set to 1 as a placeholder to prevent errors (e.g. ``divide by zero" errors) and the user will be flagged about the placeholder by a comment in the code and an error log output in the command terminal when running the program. The user can manually replace the placeholder value with its actual value.
		\item Fill in the implementation template of the chosen dynamical equation using the parameters found in the previous step. This will result in completed code snippet for the function body.
		\item Automatically edit the initialization function of the class (see \Cref{sec:skeleton_gen})  to include the states of the equation that has been generated in the function body. This can be done automatically since the states of the dynamical equation can be inferred from the implementation template chosen.
		\item Automatically generate a ``simulate" function that calls all the functions with generated function bodies.
        
        \item Recursively add a ``simulate" function to each composite parent class of the current class until a top-level component class is reached. This function should call the composite child class's ``simulate" function and return its output. This ``simulate" function is used during the computational model stage to propagate the output of the states of each component to the computational model. There are no user inputs at this stage. 
		\item Add the missing imports to the top of the file. The imports necessary are inferred from the dynamical equation chosen to be implemented in the function body. (e.g. if the dynamical equation chosen requires numpy, numpy will be added to the imports)
	\end{enumerate}

	An example of the implemented function and its augmented class from its original structure in \Cref{lst:Motion_class_example} is shown in \Cref{lst:example_code_class}. It is possible to observe that the appropriate dynamical equation template has been selected and that the parameters necessary to execute the function have been accurately captured. 

	\subsection{Computational model generation}
	\label{sec:simulation_generation_methodology}
	When having a working code that models each of the separate components of the dynamical system, a computational model is obtained by combining these components to produce a simulation of the system that is necessary to draw conclusions about the performance of the system.  This can be done by creating a highest level system class that initialises the class instances of all the components of the system, sets all the initial conditions and simulates each of them at each timestep. An example system file can be seen in \Cref{lst:example_system_class}.  The simulation results (i.e., the values of the states of the dynamical system) can then be used to compute performance metrics for the system or to verify whether the states exceed safe values for a system's operation.
    
    The generation of the system class is done by initialising all of the top-level components that subsequently initialise all their child components in an initialization function (the function called during the construction of a class instance).   A class function named ``simulate" is then created for the system class. This ``simulate" function takes in simulation parameters and initial conditions as user input. Simulation parameters would include the length of each timestep and the total time of the simulation. The required initial conditions for user input depend on the dynamical equation being simulated. The outputs of these simulate functions will be the states of each component in the dynamical system at each timestep. These are be compiled into arrays of states at each timestep forming the output of the simulation. 

    If any of the required inputs for the user is not present, the input value will be instantiated to predefined a default value. This function will iteratively call the ``simulate" function at of each top-level class that has their own ``simulate" function (as described in \Cref{sec:genBodyStrat}) at each timestep. 
    
    The simulation output data is hard to analyse purely as an array of numbers representing the states of the system at each timestep. Therefore, the data can be better analyzed by plotting them into a graph or by assessing some specific performance metrics.   The template for the system class can be found in \Cref{lst:example_system_class_template}.

\section{Validation of proposed approach: from unstructured text on a simple dynamical system to the computational model}\label{sec:validation}

 The proposed approach from text to computational model was validated by performing an end-to-end case study of the approach applied to a simple pendulum system described in \Cref{fig:simple_pendulum_example}.

\begin{figure}[h!]
		\centering
		\fbox{
			\begin{minipage}{\textwidth}
				\begin{center}
					\textbf{Source text:}\\
					 A pendulum is made up of a bob attached to an inextensible string fixed at one end. The bob weighs 2 kilograms. The string is 1.5 meters long. Gravity causes the pendulum to swing. The motion is periodic and can be approximated by simple harmonic motion for small angles. The pendulum's period depends on its length and gravity. The mass of the bob does not affect the period.
				\end{center}
			\end{minipage}
		}
		\caption{Example input text for a simple pendulum system.}
		\label{fig:simple_pendulum_example}
	\end{figure}

	\subsection{BDD Diagram Generation}

	Using the text in \Cref{fig:simple_pendulum_example}, and a collection of patents as the corpus, as inputs into the steps described in \Cref{sec:prep}, \Cref{sec:Kgrap} and \Cref{sec:BDDGen}, A BDD diagram that captures the components of the system and their relationships was generated. The generated BDD diagram is shown in \Cref{fig:simple_pendulum_bdd}. The partial outputs up to the key relationships for the proposed approach are shown in \Cref{tab:pendulum_extraction_partial outputs}, while the extracted attributes for the simple pendulum system are shown in \Cref{tab:pendulum_attrs}.





\begin{table}[h!]
	\centering
	\begin{tabular}{|p{4cm}|p{9cm}|}
	\hline
	\textbf{Partial Outputs} & \textbf{Values} \\
	\hline
	Key Nouns & Pendulum, bob, mass, gravity, meter, oscillates, resembles, period, motion, kilogram, string, cause, move, angle, end \\
	\hline
	Extracted Relationships & 
	\begin{tabular}[c]{@{}l@{}}
	(An inextensible string, fixed, at one end) \\
	(A bob, attached, ) \\
	(A pendulum, is made up, of a bob) \\
	(The bob, weighs, 2 kilograms) \\
	(The string, has length of, 1.5 meters) \\
	(The string, is, 1.5 meters long) \\
	(The pendulum, to swing, ) \\
	(Gravity, causes, the pendulum to swing) \\
	(The motion, can be approximated, by simple harmonic motion \\ for small angles) \\
	(The motion, is, periodic) \\
	(The pendulum's period, depends, on gravity) \\
	(The pendulum's period, depends, on its length) \\
	(The mass of the bob, does not affect, the period)
	\end{tabular} \\
	\hline
	Key Phrases & pendulum, bob, mass bob, pendulum period, period, kilogram, gravity, end, string, motion, motion angle, meter \\
	\hline
	Key Relationships & \begin{tabular}[c]{@{}l@{}}
    (string, fixed, end)\\
    (pendulum, is made up, bob)\\
    (bob, weighs, kilogram)\\
    (string, has length of, meter)\\
    (string, is, meter)\\
    (gravity, causes, pendulum)\\
    (motion, can be approximated, motion angle)\\
    (pendulum period, depends, gravity)\\
    (mass bob, does not affect, period)\end{tabular} \\
	\hline
	\end{tabular}
	\caption{Summary of partial outputs of the proposed approach up to key relationships for the simple pendulum system.}
	\label{tab:pendulum_extraction_partial outputs}
	\end{table}

    \begin{table}[h!]
        \centering
        \begin{tabular}{|c|c|}
            \hline
            \textbf{Block} & \textbf{Extracted Attributes} \\
            \hline
            pendulum period & gravity: gravity, period: s, length: period \\
            \hline
            bob & weight: 2kg \\
            \hline
            pendulum & fixed: at one end, string: inextensible,
            bob: attached \\
            \hline
            motion & type: periodic,
            approximation: simple harmonic, condition: small angles \\
            \hline
            string & string: inextensible, length: 1.5 meters, bob: attached, fixed: one end \\
            \hline
        \end{tabular}
        \caption{Automated extracted attributes for blocks for simple pendulum system}
        \label{tab:pendulum_attrs}
    \end{table}

    The key phrase extraction validation results for the simple pendulum system are shown in \Cref{tab:key_phrase_extraction_validation_simple_pendulum}. It is shown that the recall in both exact match and fuzzy match was very good, meaning that all key phrases in the ground truth are extracted, however the precision is low in both the exact match and the fuzzy match cases, demonstrating that many extracted phrases were not present in the ground truth. This is also visible from the normalized extracted phrase number showing that the proposed strategy extracted 4 times the number of key phrases present in the ground truth. It is very difficult to reduce the amount of extracted key phrases as it might remove an essential key phrase and compromise the recall. The recall is more significant than the precision for our purposes.

    \begin{table}
	\centering
	\begin{tabular}{|l|c|}
	\hline
	\textbf{Metric} & \textbf{Value} \\
	\hline
	Exact Match Recall & 1.00 \\ \hline
	Exact Match Precision & 0.25 \\\hline
	Fuzzy Match Recall & 1.00 \\\hline
	Fuzzy Match Precision & 0.25 \\\hline
	Normalized Extracted Phrase Number & 4 \\
	\hline
	\end{tabular}
	\caption{Key Phrase Extraction Validation Metrics.}
	\label{tab:key_phrase_extraction_validation_simple_pendulum}
	\end{table}
	
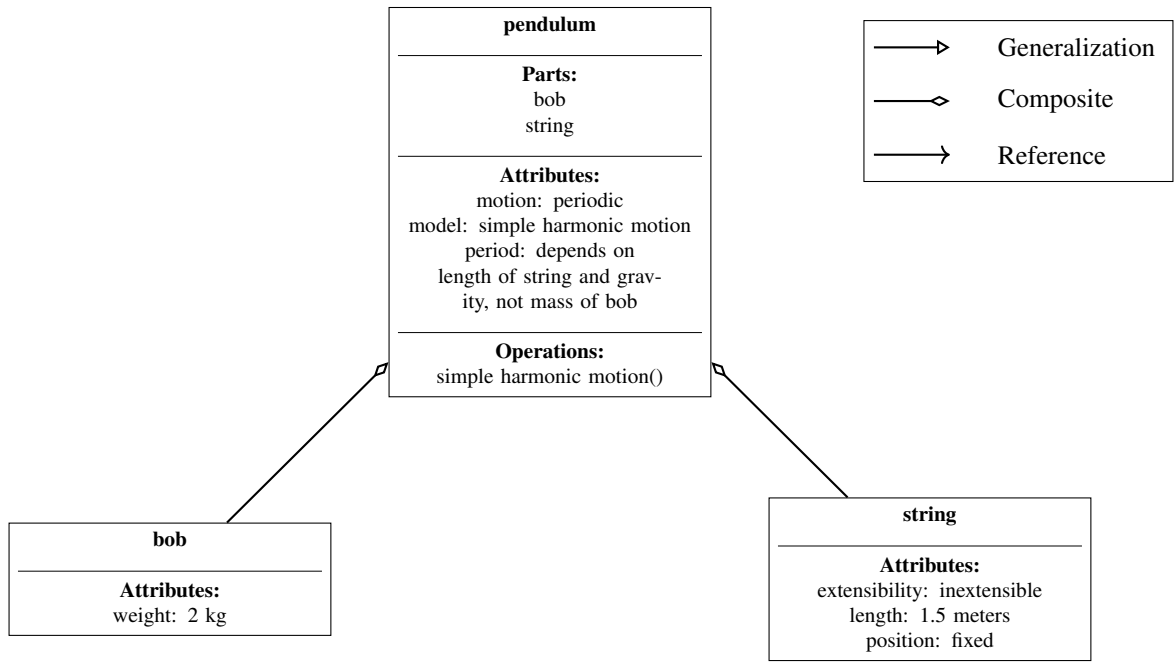
\begin{figure}
    \centering
    \begin{tikzpicture}[
        block/.style={rectangle, draw, text width=4cm, text centered, font=\footnotesize},
		abstractblock/.style={rectangle, draw, text width=4cm, text centered, font=\footnotesize, dashed},
        generalization/.style={->, thick, >={Triangle[open]}},
        composite/.style={->, thick, >={Diamond[open]}},
        reference/.style={->, thick},
        abstraction/.style={->, thick, dotted}
    ]

	\node[block] (bob) at (-5, -10) {%
		\textbf{bob} \\ \rule{\linewidth}{0.3pt}
		\textbf{Attributes:} \\
		weight: 2 kg \\

	};

	\node[block] (pendulum) at (0, -5) {%
		\textbf{pendulum} \\ \rule{\linewidth}{0.3pt}
		\textbf{Parts:} \\
		bob \\
		string \\ \rule{\linewidth}{0.3pt}
		\textbf{Attributes:} \\
		motion: periodic\\
		model: simple harmonic motion \\
		period: depends on length of string and gravity, not mass of bob \rule{\linewidth}{0.3pt}
		\textbf{Operations:} \\
		simple harmonic motion()
	};

	\node[block] (string) at (5, -10) {%
		\textbf{string} \\ \rule{\linewidth}{0.3pt}
		\textbf{Attributes:} \\
		extensibility: inextensible \\ 
		length: 1.5 meters \\ 
		position: fixed
	};

	\matrix [draw, below right, xshift=2cm, yshift=-1cm, column sep=0.5cm, row sep=0.2cm] at (pendulum_period.south east) {
		\draw[generalization] (0,0) -- +(1,0); & \node[anchor=west] {Generalization}; \\
		\draw[composite] (0,0) -- +(1,0); & \node[anchor=west] {Composite}; \\
		\draw[reference] (0,0) -- +(1,0); & \node[anchor=west] {Reference}; \\
	};

	\draw[composite] (bob) -- (pendulum);
	\draw[composite] (string) -- (pendulum);

    \end{tikzpicture}
    \caption{ Ground Truth Block Definition Diagram for the Simple Pendulum System.}
    \label{fig:simple_pendulum_bdd_gt}
\end{figure}

	A manual validation of the BDD diagram shown in \Cref{fig:simple_pendulum_bdd} can be done by comparing it with \Cref{fig:simple_pendulum_bdd_gt}. The BDD diagram captures the core pendulum's components in the ground truth, such as the bob, string and pendulum. If the user does not remove the non-essential key phrases in the key phrase extraction step, there will be many unrelated components extracted in the generated BDD diagram. This is not critical as it does not prevent the model from being generated as shown in the next section. For this validation case, the non-essential key phrases were not manually removed to validate the performance of the algorithm without manual intervention. The generated BDD diagram also captures the relationships between the components, such as the composition relationship between the pendulum and the bob, and the generalization relationship between the pendulum period and the period.

    \begin{table}
	\centering
	\begin{tabular}{|l|c|}
	\hline
	\textbf{Similarity/match score} & \textbf{Score} \\ \hline
    Block Similarity: bob - bob & 1.00 \\\hline
	Block Similarity: string - string & 0.67 \\ \hline
    Block Similarity: pendulum - pendulum & 0.00 \\ \hline
    Set Similarity & 0.56 \\ \hline
    Zero Attribute Match Score & 1.00 \\ \hline
	Normalized Set Match Score & 1.00 \\ \hline
	\end{tabular}
	\caption{Results of Attribute value extraction validation for the simple pendulum system.}
	\label{tab:pendulum_block_attr_validation}
	\end{table}

	\begin{figure}[h!]
    \centering
    \begin{tikzpicture}[
        block/.style={rectangle, draw, text width=4cm, text centered, font=\footnotesize},
		abstractblock/.style={rectangle, draw, text width=4cm, text centered, font=\footnotesize, dashed},
        generalization/.style={->, thick, >={Triangle[open]}},
        composite/.style={->, thick, >={Diamond[open]}},
        reference/.style={->, thick},
        abstraction/.style={->, thick, dotted}
    ]

	\node[block] (pendulum_period) at (0, 0) {%
		\textbf{pendulum period} \\ \rule{\linewidth}{0.3pt}
		\textbf{Attributes:} \\
		gravity: gravity \\ 
		period: s \\ 
		length: period \\ \rule{\linewidth}{0.3pt}
		\textbf{Operations:} \\
		depends()
	};

	\node[block] (period) at (-5, 2) {period};

	\node[block] (mass_bob) at (-5, -1) {%
		\textbf{mass bob} \\ \rule{\linewidth}{0.3pt}
		\textbf{Parts:} \\
		period \\ \rule{\linewidth}{0.3pt}
		\textbf{Operations:} \\
		does not affect()
	};

	\node[block] (bob) at (-5, -5) {%
		\textbf{bob} \\ \rule{\linewidth}{0.3pt}
		\textbf{Attributes:} \\
		weight: 2 kg \\ \rule{\linewidth}{0.3pt}
		\textbf{Operations:} \\
		weighs()
	};

	\node[block] (pendulum) at (0, -5) {%
		\textbf{pendulum} \\ \rule{\linewidth}{0.3pt}
		\textbf{Attributes:} \\
		fixed: at one end \\ 
		string: inextensible \\ 
		bob: attached \\ \rule{\linewidth}{0.3pt}
		\textbf{Operations:} \\
		is made up()
	};

	\node[block] (motion) at (-5, -13) {%
		\textbf{motion} \\ \rule{\linewidth}{0.3pt}
		\textbf{Attributes:} \\
		type: periodic \\ 
		approximation: simple harmonic \\ 
		condition: small angles \\ \rule{\linewidth}{0.3pt}
		\textbf{Operations:} \\
		can be approximated()
	};

	\node[block] (motion_angle) at (-5, -17) {motion angle};

	\node[block] (gravity) at (0, -9) {%
		\textbf{gravity} \\ \rule{\linewidth}{0.3pt}
		\textbf{Parts:} \\
		pendulum \\ \rule{\linewidth}{0.3pt}
		\textbf{Operations:} \\
		causes()
	};

	\node[block] (string) at (0, -16) {%
		\textbf{string} \\ \rule{\linewidth}{0.3pt}
		\textbf{Attributes:} \\
		string: inextensible \\ 
		length: 1.5 meters \\ 
		bob: attached \\ 
		fixed: one end \\ \rule{\linewidth}{0.3pt}
		\textbf{Parts:} \\
		meter \\ \rule{\linewidth}{0.3pt}
		\textbf{Operations:} \\
		has length() \\ 
		fixed() \\ 
		is()
	};

	\node[abstractblock] (physical_entity) at (6, -8) {physical entity};

	\node[block] (end) at (6, -12) {end};

	\node[abstractblock] (object) at (6, -13) {object};

	\node[block] (meter) at (6, -15) {meter};

	\node[block] (kilogram) at (-5, -8) {%
		\textbf{kilogram} \\ \rule{\linewidth}{0.3pt}
		\textbf{Parts:} \\
		bob
	};

	\node[abstractblock] (entity) at (0,-11.5) {entity};

	\node[abstractblock] (abstraction) at (-5, -10) {abstraction};

	\matrix [draw, below right, xshift=2cm, yshift=-1cm, column sep=0.5cm, row sep=0.2cm] at (pendulum_period.south east) {
		\draw[generalization] (0,0) -- +(1,0); & \node[anchor=west] {Generalization}; \\
		\draw[composite] (0,0) -- +(1,0); & \node[anchor=west] {Composite}; \\
		\draw[reference] (0,0) -- +(1,0); & \node[anchor=west] {Reference}; \\
	};

	\draw[generalization] (pendulum_period) -- (period);
	\draw[composite] (period) -- (mass_bob);
	\draw[generalization] (mass_bob) -- (bob);
	\draw[composite] (bob) -- (kilogram);
	\draw[reference] (bob) -- (pendulum);
	\draw[composite] (motion_angle) -- (motion);
	\draw[composite] (gravity) -- (pendulum);
	\draw[reference] (end) -- (string);
	\draw[composite] (meter) -- (string);
	\draw[generalization] (motion) -- (abstraction);
	\draw[generalization] (kilogram) -- (abstraction);
	\draw[generalization] (gravity) -- (entity);
	\draw[generalization] (motion) -- (entity);
	\draw[generalization] (kilogram) -- (entity);
	\draw[generalization] (string) -- (entity);
	\draw[generalization] (end) -- (entity);
	\draw[generalization] (string) -- (physical_entity);
	\draw[generalization] (gravity) -- (physical_entity);
	\draw[generalization] (end) -- (physical_entity);
	\draw[generalization] (end) -- (object);
	\draw[generalization] (string) -- (object);

    \end{tikzpicture}
    \caption{Block Definition Diagram for the Simple Pendulum System.}
    \label{fig:simple_pendulum_bdd}
\end{figure}

\subsection{Code Generation}
\begin{figure}
		\centering
		{\footnotesize
		\begin{tikzpicture}[x=1em, y=1.2em, every node/.style={anchor=west}]
			\node (root) at (0,0) {\faFolderOpen~\texttt{Simple Pendulum System}};
			\node (gravitypy) at (2,-1) {\faFile~\texttt{Gravity.py}};
			\node (kilogrampy) at (2,-2) {\faFile~\texttt{Kilogram.py}};
			\node (motionpy) at (2,-3) {\faFile~\texttt{Motion.py}};
			\node (motionanglepy) at (2,-4) {\faFile~\texttt{MotionAngle.py}};
			\node (stringpy) at (2,-5) {\faFile~\texttt{String.py}};
			\node (systempy) at (2,-6) {\faFile~\texttt{System.py}};
			\node (gravityparts) at (2,-7) {\faFolderOpen~\texttt{Gravity\_parts/}};
			\node (kilogramparts) at (2,-9) {\faFolderOpen~\texttt{Kilogram\_parts/}};
			\node (stringparts) at (2,-15) {\faFolderOpen~\texttt{String\_parts/}};
			\node (pendulum) at (5,-8) {\faFile~\texttt{Pendulum.py}};
			\node (massbob) at (5,-10) {\faFile~\texttt{MassBob.py}};
			\node (bob) at (5,-11) {\faFile~\texttt{Bob.py}};
			\node (massbobparts) at (5,-12) {\faFolderOpen~\texttt{MassBob\_parts/}};
			\node (pendulumperiod) at (8,-13) {\faFile~\texttt{PendulumPeriod.py}};
			\node (period) at (8,-14) {\faFile~\texttt{Period.py}};
			\node (endpy) at (5,-16) {\faFile~\texttt{End.py}};
			\node (meterpy) at (5,-17) {\faFile~\texttt{Meter.py}};
			\draw[-] (0.5,-0.5) -- (0.5,-1) -- (2,-1);
			\draw[-] (0.5,-0.5) -- (0.5,-2) -- (2,-2);
			\draw[-] (0.5,-0.5) -- (0.5,-3) -- (2,-3);
			\draw[-] (0.5,-0.5) -- (0.5,-4) -- (2,-4);
			\draw[-] (0.5,-0.5) -- (0.5,-5) -- (2,-5);
			\draw[-] (0.5,-0.5) -- (0.5,-6) -- (2,-6);
			\draw[-] (0.5,-0.5) -- (0.5,-7) -- (2,-7);
			\draw[-] (0.5,-0.5) -- (0.5,-9) -- (2,-9);
			\draw[-] (0.5,-0.5) -- (0.5,-15) -- (2,-15);
			\draw[-] (3,-7.5) -- (3,-8) -- (5,-8);
			\draw[-] (3,-9.5) -- (3,-10) -- (5,-10);
			\draw[-] (3,-9.5) -- (3,-11) -- (5,-11);
			\draw[-] (3,-9.5) -- (3,-12) -- (5,-12);
			\draw[-] (6,-12.5) -- (6,-13) -- (8,-13);
			\draw[-] (6,-12.5) -- (6,-14) -- (8,-14);
			\draw[-] (3,-15.5) -- (3,-16) -- (5,-16);
			\draw[-] (3,-15.5) -- (3,-17) -- (5,-17);
		\end{tikzpicture}
		}
		\caption{File structure generated from the BDD diagram for the simple pendulum system.}
		\label{fig:pendulum_folder_structure}
	\end{figure}
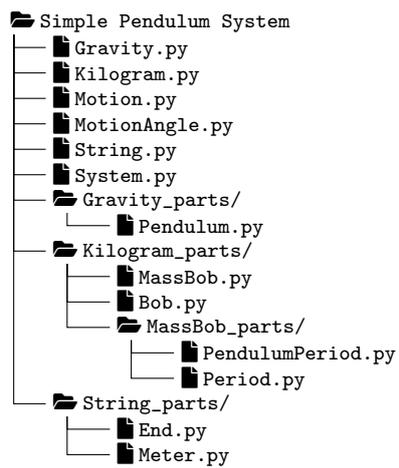

The computational model's code was generated using the generated BDD diagram extracted according to the approach in \Cref{sec:code_generation_methodology}. The file structure of the code is shown in \Cref{fig:pendulum_folder_structure}. The file structure generated clearly follows the hierarchy of the BDD diagram, with each block having its own file and subdirectories for composite blocks. 

The class skeleton generated for the motion block is shown in \Cref{lst:Motion_class_example} and its completed class with the function bodies generated is shown in \Cref{lst:example_code_class}. The ``CanBeApproximated" function clearly implements the dynamical equation for the simple pendulum system, with the parameters being set to the values extracted from the BDD diagram. The code is structured in a way that is easy to understand and modify, with clear comments for each class and function.

\subsection{Computational model generation}
\label{sec:case_sim_gen}
\begin{figure}
		\centering
		\includegraphics[width = \textwidth]{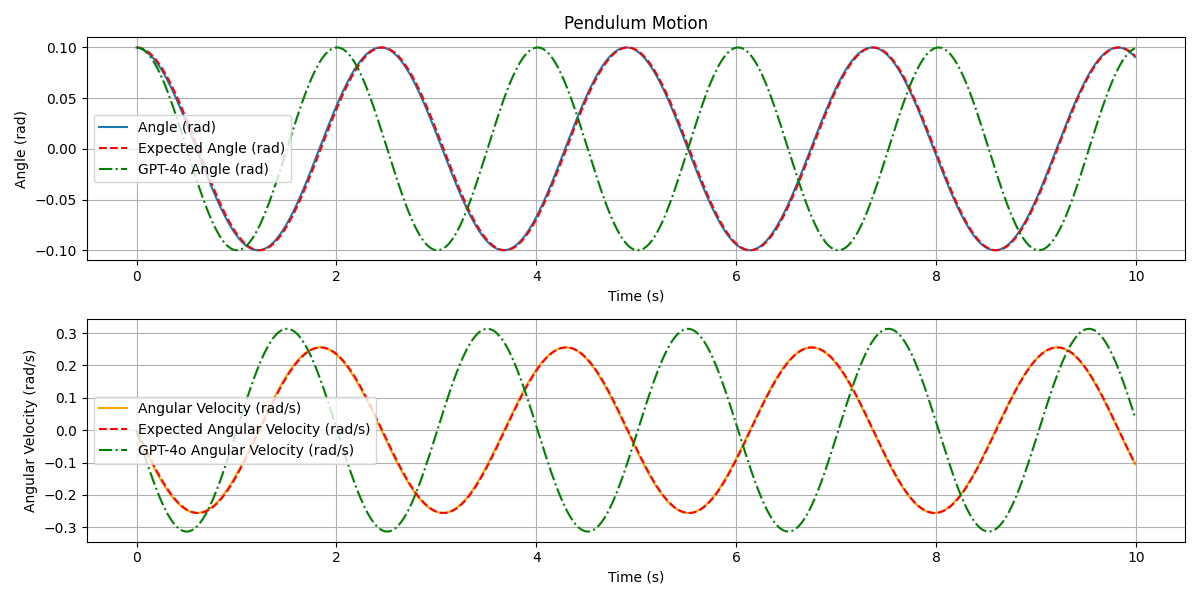}
		\caption{Simulation results for the generated code for the simple pendulum system. The plot also shows the simulation results for the code generated by Copilot (GPT-4o) given the code structure for the dynamical system.}
		\label{fig:pendulum_sim_results}
	\end{figure}

System class file was automatically generated for the simple pendulum system as shown in \Cref{lst:example_system_class}, where the computational model
is generated by initialising all of the top-level components of the pendulum system. The engineer can provide simulation parameters and the initial angle to run the simulation. The simulation was run with the engineer specifying a 0.1 rad initial angle, 1 second timesteps and a 100 second long simulation time. The system file then calls the ``simulate" function from the ``Motion" class repeatedly to obtain the states (the angle and angular velocity of the pendulum) of the system at every timestep. The simulation results are plotted and shown in \Cref{fig:pendulum_sim_results}, which shows the expected periodic motion of the pendulum.

For validation, the Github copilot (using GPT-4o) was fed the file structure with class skeletons (without the function bodies completed) to implement the function bodies and generate the simulation. The prompting and code generated can be found at \url{https://github.com/Idealistmatthew/IIBProj_Text2CompModel/blob/main/Notebooks/copilot_completion.md}, whilst the simulation results can be seen at \Cref{fig:pendulum_sim_results}. The results show that the the code generated by Copilot did successfully infer the correct equations for the system, however it hallucinated the values for the length of the string and the mass of the bob to be 1m and 1kg respectively. This shows that the proposed approach
has the potential to generate code given the code structure that is more accurate than the code generated by Copilot that is affected by  hallucinations.

\section{Limitations of the proposed work and possible extensions}\label{sec:discussion}

The results presented demonstrate that the proposed approach is capable of automatically generating the computational model of a dynamical system directly from unstructured natural language text by exploiting System modeling Language (SysML) diagrams and by making strategic use of NLP strategies and very limited use of LLMs.  By investigating BDD diagram results for different input texts, it was shown the advantage of the proposed approach over zero-shot LLM prompting. It was demonstrated that the generated BDD diagrams can sufficiently accurately represent the structure and components of dynamical systems and their relations. Furthermore, new validation strategies for blocks with attributes were proposed, providing a foundation for future research in this area and enabling the engineer to identify steps which might require modifications. 

The proposed approach is a stepping stone towards: (i) AI-powered design-space exploration to accelerate concept ideation and product realization; (ii) enhancing information sharing across teams and team-level workflow for dynamical systems designs by integrating domain knowledge and simulation environments while keeping the human-in-the-loop for critical decision making at every step of the approach; (iii) accelerating Digital Twin \cite{Tsialiamanis2021} development and deployment; (iv) automatically extracting domain and expert knowledge in a format that can be readily integrated with real-world data and physics-based models within Physics-Enhanced Machine Learning strategies \cite{Cicirello2024}. 
Possible extensions of the current work include:
\begin{itemize}
       \item \textbf{Interactive tool development}: The proposed approach has been applied end-to-end as a proof-of-concept. It could be embedded in an interactive tool to generate domain-specific languages that can be directly integrated with dedicated engineering  software tools (to automatically build, run and visualize the results of the dynamical system computational model), while enabling the human-in-the-loop to provide feedback to the algorithm at each step as to what constitutes a good or bad overall result considering information which might have not be possible to be captured from the natural language text only. Such a tool could also allow for more explicit calibration by the engineer or by optimization algorithms of the various parameters that need to be specified with the proposed approach and within the computational model;
       \item \textbf{Interfaces between components}: The current approach does not explicitly fully capture the complexity of the interfaces between the components of a complex built-up dynamical systems (e.g., nonlinear joints behavior) and their interactions. This is a crucial part for the development of accurate computational models, and currently it would require manual input from an engineer. Future work should focus on extracting and modeling interface information between components;

         \item \textbf{Accurate representation of causal information}: The goal of the generated computational model is to provide a physics-based relationship between cause and effect that can be used to explore the behavior of the system. Therefore, the identification of causal sentences within natural language text is another critical aspect for the development of complex dynamical systems computational models. Currently, the causal information is extracted in the relationship extraction stage. This process might be non-exhaustive, leading to potential loss of information, and insufficiently descriptive. For example, the sentence ``the cantilever has a fixed end" would correspond to a relationships of the form (the cantilever, has, a fixed end) instead of explicitly identifying the physical meaning  describing the boundary condition.  As discussed in \cite{causality}, future work should focus on developing tools for distinguishing from natural language text information regarded as causal to the dynamical system behavior (e.g., loading conditions) from information relevant for describing the dynamical system components. Moreover, challenges \cite{causality} such as  implicit causality, embedded causality and negative causality remain to be addressed. Further, once the natural language text information regarded as causal to the dynamical system behavior is extracted, additional steps might be needed to fully map this information accurately in the computational model; 
         
    \item \textbf{Multimodal Data Processing}: Engineering documents are often multimodal, containing critical information about the dynamical system in text, figures, tables, diagrams and equations. For example, the patent  investigated in \Cref{sec:observe_BDD} had also context-rich figures. However, the current strategy  focuses only on extraction and processing of information from natural language text. Future work should explore integrating information extraction from image for improving the generated knowledge diagram. Moreover,  extracting terms/equations to be inputted in the code generation process would enable the approach to handle a wider range of engineering documents and improve the accuracy of the generated models.  This is of particular relevant for complex multi-physics and multi-scale problems whose description from text only might be limited;
\end{itemize}

It is worth mentioning that comparison with LLM were performed with zero-shot prompting, and alternative comparisons could be have carried out by providing  the LLM model with examples of correct descriptions (e.g., in the case of diagrams) or the required code. This is considered beyond the scope of the current work. 

\section{Conclusions}\label{sec:conclusions}

The work presented in this paper demonstrates the viability and advantages of non-LLM-centric approaches for generating dynamical systems (deterministic) computational models from unstructured natural language text by exploiting SysML diagrams. The key advancements in the proposed approach are the enhanced automated SysML diagram generation, its validation,  and the integration of SysML with tools for enabling an end-to-end process from text to model.  The results obtained with the proposed approach in terms of SysML diagrams were validated against manually generated and ChatGPT generated BDDs, showing that the proposed approach is context-free, capable of handling diverse document types even with a limited information, and that it can be extended to more complex systems. Moreover, an end-to-end case study on a simple pendulum has shown the applicability of the proposed approach for generating computational models of dynamical systems starting from a short text description. Rather than replacing engineering personnel, the proposed approach is intended to support them by providing an initial SysML diagram and its corresponding computational model that can be further refined and developed.  This is achieved by providing intermediate outputs that can be controlled by the engineering team. Moreover, each step in this approach is not implementation-specific, and it could be carried out by exploiting alternative tools or algorithms. 

The key enabler of the proposed approach has been the enhancement of the automated SysML diagrams initially proposed in \cite{zhong2023natural} by introducing: (i) a spelling correction preprocessing step to improve component name recognition; (ii) coreference resolution to ensure consistent mapping of entities within BDD diagrams; and (iii) an automatic extraction of component attributes strategy for capturing essential information for modeling each component (e.g. information on the properties, behaviors and constraints of each component in the system). 
Moreover, a strategy for automatically generating an object-oriented computational model whose structure reflects that of the underlying dynamical system starting from the generated SysML diagram has been proposed. The first step consists of converting the BDD diagram into code that models the components of the dynamical system. By leveraging domain and expert knowledge (in the form of equation implementation templates), summarization tools and context extraction, informative docstrings for classes and functions were generated. The proposed computational model generation step uses the generated code to run simulations of the dynamical system across various input conditions. 
\\
\textbf{Data Availability Statement}	\label{sec:github_repo}\\
The code and data needed to replicate the results shown in this paper can be accessed in the GitHub repository at \url{https://github.com/Idealistmatthew/IIBProj_Text2CompModel}
\\
\textbf{Acknowledgments}
\\ 
The authors would like to thank Shaohong Zhong for helpful discussions at the beginning of this project. 
\bibliography{mybibfile}

\appendix
\section{Documents Used in Paper}

\begin{table}
		\centering
		\begin{tabular}{|p{3cm}|p{5cm}|p{3cm}|p{4cm}|}
			\hline
			\textbf{Input document} & \textbf{Example Excerpt} & \textbf{Corpus Used} & \textbf{Usage Motivation} \\
			\hline
			Chapter 7 from Jackman, W.J. (1912) ``Flying Machines: Construction and Operation" \cite{jackman1912flying} & 
		``The rudder beams form the top and bottom frames of the vertical rudder. To these are bolted and clamped two upright pieces, 3 feet, 10 inches in length, and 3/4 inch in cross section ..." & 
			All other chapters from the same book (27 chapters in total) & 
			Rich, domain-specific language and complex system descriptions; tests extraction on historical engineering text \\
			\hline
			Patent JP6875871B2 \cite{yasuda2021jp6875871b2} & 
			``A plurality of exhaust valve cams 46a are fixed to the exhaust camshaft 46. The exhaust valve cam 46a is in contact with the end of the exhaust valve 40, and rotates with the rotation of the exhaust cam shaft 46 to move the exhaust valve 40 in the axial direction." & 
			A collection of 10 engineering system patents & 
			Modern technical language, formal structure; evaluates method on contemporary engineering documents \\
			\hline
			Simple Pendulum System Description (Synthetic) (Full text is only 68 words long) & 
			``A simple pendulum consists of a mass attached to a string, which swings under the influence of gravity." & 
			A collection of 10 engineering system patents & 
			Baseline for validation, tests pipeline on simple, well-understood systems \\
			\hline
		\end{tabular}
		\caption{Summary of documents used for testing, with example excerpts, corpus selection, and usage motivation.}
		\label{tab:document_types}
	\end{table}

\section{Coreference Resolution Example}
	\begin{figure}[H]
		\centering
		\fbox{
			\begin{minipage}{0.8\textwidth}
				\begin{center}
					\textbf{Sentence before Coreference Resolution:} A \textcolor{darkgreen}{screw propeller} working under load approaches more closely to \textcolor{red}{its} maximum efficiency as it carries \textcolor{red}{its} load with a minimum amount of slip, or nearing \textcolor{red}{its} calculated pitch speed. \\
									\begin{center}
										\begin{tikzpicture}
											\draw[->, thick] (0,0) -- (0,-1);
										\end{tikzpicture}
									\end{center}
					\textbf{Sentence after Coreference Resolution:} A \textcolor{darkgreen}{screw propeller} working under load approaches more closely to \textcolor{darkgreen}{propeller} maximum efficiency as \textcolor{darkgreen}{propeller} carries \textcolor{darkgreen}{propeller} load with a minimum amount of slip , or nearing \textcolor{darkgreen}{propeller} calculated pitch speed .
				\end{center}
			\end{minipage}
		}
		\caption{Coreference Resolution applied to an excerpt from the book \cite{jackman1912flying}. The text in red are implicit mentions of the screw propeller, while the text in green are the explicit mentions of the screw propeller.}
		\label{fig:coref_resolution_example}
	\end{figure}
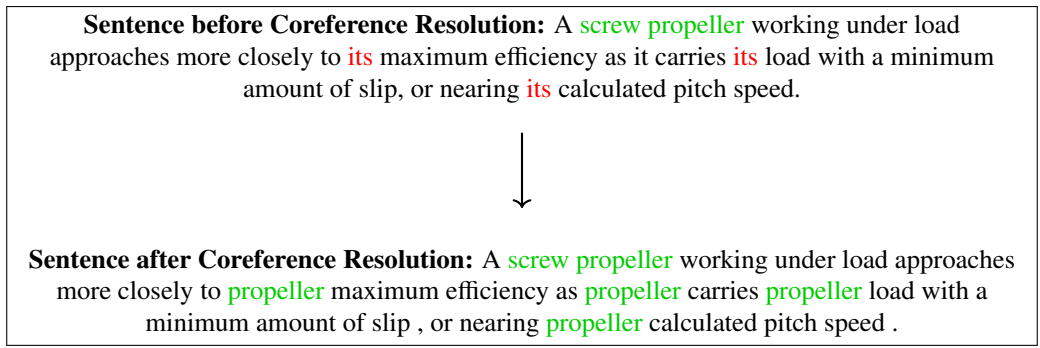

    \section{Penn Treebank Tagset}

    \begin{table}
    \centering
    \begin{tabular}{llp{8cm}}
    \hline
    \textbf{Tag} & \textbf{POS} & \textbf{Examples} \\
    \hline
    DT   & Determiner & the, a, some, most, every, etc. \\
    NN   & Noun, singular or mass & cat, school, milk \\
    VBG  & Verb, gerund/present participle & running, eating \\
    IN   & Preposition or subordinating conjunction & in, of, like, because, although \\
    VBZ  & Verb, 3rd person singular present & runs, eats \\
    RBR  & Adverb, comparative & faster, better \\
    RB   & Adverb & quickly, silently, well \\
    TO   & to & infinitival "to" \\
    PRP  & Personal pronoun & I, you, he, she, it \\
    JJ   & Adjective & green, quick, absurd, etc. \\
    \hline
    \end{tabular}
    \caption{Selected Penn Treebank POS Tags \cite{bird2009natural}}
    \label{tab:ptb-selected-tags}
    \end{table}

    \section{Relationship Extraction Example}
    \begin{figure}[H]
		\centering
		\fbox{
			\begin{minipage}{0.8\textwidth}
				\begin{center}
					\textbf{Example Sentence:} The real motor-propelled flying machine, generally has both front and rear rudders manipulated by wire cables at the will of the operator.
					\vspace{-0.5em}
									\begin{center}
										\begin{tikzpicture}
											\draw[->, thick] (0,0) -- (0,-0.5);
										\end{tikzpicture}
									\end{center}
									\vspace{-0.5em}
					\textbf{Extracted Relationships:} \\
					(motor-propelled flying machine, has, front rudder) \\
					(motor-propelled flying machine, has, rear rudder) \\
					(motor-propelled flying machine, manipulated by, wire cables) \\
					(wire cables, manipulated by, operator) \\
				\end{center}
			\end{minipage}
		}
		\caption{Relationship extraction applied to an excerpt from the book \cite{jackman1912flying}.}
		\label{fig:rel_extraction_example}
	\end{figure}
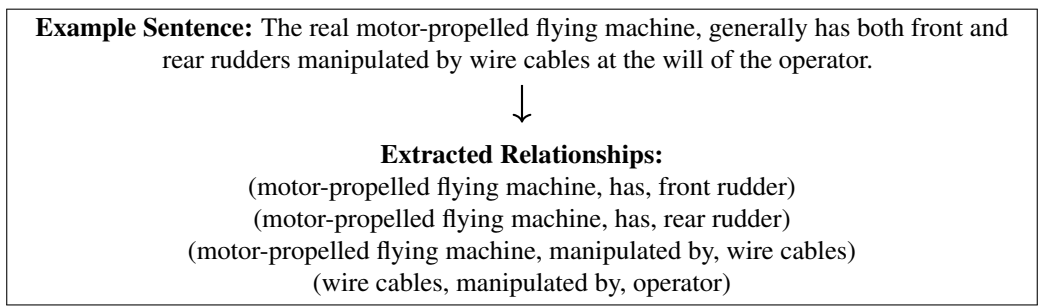

    \section{Knowledge Graph Example}
    \begin{figure}[H]
		\centering
		\begin{tikzpicture}[node distance=3cm, auto]
			\node[draw, rectangle] (flyingMachine) {Motor-Propelled Flying Machine};
			\node[draw, rectangle, below left of=flyingMachine] (frontRudder) {Front Rudder};
			\node[draw, rectangle, below right of=flyingMachine] (rearRudder) {Rear Rudder};
			\node[draw, rectangle, above left of=flyingMachine] (wireCables) {Wire Cables};
			\node[draw, rectangle, right of=wireCables, node distance=7cm] (operator) {Operator};

			\draw[->] (flyingMachine) -- (frontRudder) node[midway, above left] {has};
			\draw[->] (flyingMachine) -- (rearRudder) node[midway, above right] {has};
			\draw[->] (flyingMachine) -- (wireCables) node[midway, above right] {manipulated by};
			\draw[->] (wireCables) -- (operator) node[midway, above] {manipulated by};
		\end{tikzpicture}
		\caption{Knowledge graph made using relationships extracted in \Cref{fig:rel_extraction_example} assuming all relationships extracted were retained as key relationships.}
		\label{fig:graph_relationships}
	\end{figure}
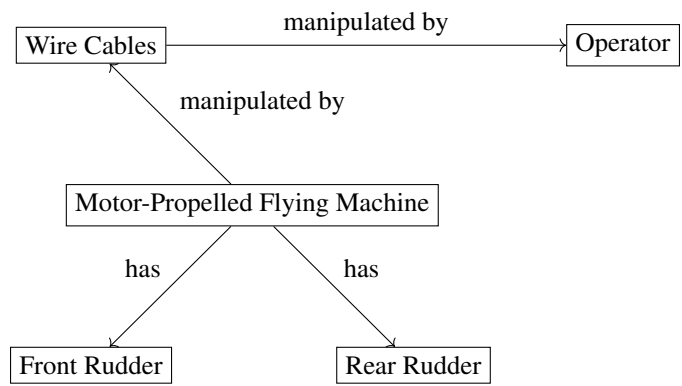

    \section{BDD Diagrams}
    \begin{figure}
    \centering
    \begin{tikzpicture}[
        block/.style={rectangle, draw, text width=4cm, text centered, font=\footnotesize},
		abstractblock/.style={rectangle, draw, dashed, text width=4cm, text centered, font=\footnotesize},
        generalization/.style={->, thick, >={Triangle[open]}},
        composite/.style={->, thick, >={Diamond[open]}},
		reference/.style={->, thick},
        abstraction/.style={->, thick, dotted}
    ]

			\node[block] (footinch) at (0, 0) {%
				\textbf{foot inch}
			};

			\node[block] (rudderbeam) at (0, -4) {%
				\textbf{rudder beam} \\ \rule{\linewidth}{0.3pt}
				\textbf{Attributes:} \\ 
				orientation: vertical \\ 
				length: 11 inches 8 feet \\ 
				\rule{\linewidth}{0.3pt}
				\textbf{Parts:} \\ 
				foot inch \\ 
				\rule{\linewidth}{0.3pt}
				\textbf{Operations:} \\ 
				is() \\ 
				is clamped() \\ 
				form()
			};

			\node[block] (bottomframerudder) at (0, -8.5) {%
				\textbf{bottom frame rudder} \\ \rule{\linewidth}{0.3pt}
				\textbf{Parts:} \\ 
				rudder beam
			};

			\node[block] (rudderbeambottom) at (-5, -6) {%
				\textbf{rudder beam bottom}
			};

			\node[block] (framerudder) at (5, -8) {%
				\textbf{frame rudder} \\ \rule{\linewidth}{0.3pt}
				\textbf{Parts:} \\ 
				rudder beam
			};

			\node[block] (rudder) at (0, -12) {%
				\textbf{rudder} \\ \rule{\linewidth}{0.3pt}
				\textbf{Attributes:} \\ 
				position: front and rear \\ 
				type: motor-propelled flying machine \\ 
				control: wire cables \\ 
				\rule{\linewidth}{0.3pt}
				\textbf{Operations:} \\ 
				manipulated()
			};

			\node[abstractblock] (rudderbottom) at (-5, -8) {%
				\textbf{rudder bottom}
			};

			\node[block] (rudderplane) at (5, -10) {%
				\textbf{rudder plane}
			};

			\node[block] (glider) at (0, -16) {%
					\textbf{glider} 
			};

			\node[block] (gliderrule) at (5, -14) {%
				\textbf{glider rule} \\ \rule{\linewidth}{0.3pt}
				\textbf{Attributes:} \\ 
				rudder: single \\ 
				\rule{\linewidth}{0.3pt}
				\textbf{Parts:} \\ 
				rudder \\ 
				\rule{\linewidth}{0.3pt}
				\textbf{Operations:} \\ 
				have()
			};

			\node[block] (becomeglideramateur) at (-6, -12) {%
				\textbf{become glider amateur} \\ \rule{\linewidth}{0.3pt}
				\textbf{Parts:} \\ 
				rudder \\ 
				\rule{\linewidth}{0.3pt}
				\textbf{Operations:} \\ 
				should equip()
			};

			\node[block] (rudderbeamglider) at (-5, -16) {%
				\textbf{rudder beam glider}
			};
				\node[abstractblock] (rudderglider) at (-5, -18) {%
					\textbf{rudder glider}
				};

			\matrix [draw, below right, xshift=1.5cm, yshift=-1cm, column sep=0.5cm, row sep=0.2cm] at (glider.south east) {
		\draw[generalization] (0,0) -- +(1,0); & \node[anchor=west] {Generalization}; \\
		\draw[composite] (0,0) -- +(1,0); & \node[anchor=west] {Composite}; \\
		\draw[reference] (0,0) -- +(1,0); & \node[anchor=west] {Reference}; \\
	};

			\draw[composite] (footinch) -- (rudderbeam)  ;
			\draw[composite] (rudderbeam) -- (bottomframerudder)  ;
				\draw[composite] (rudderbeam) -- (framerudder)  ;
			\draw[generalization] (rudderbeambottom) -- (rudderbottom); 
			\draw[generalization] (rudderbottom) -- (rudder)  ;
			\draw[generalization] (rudderplane) -- (rudder) ;
			\draw[composite] (rudder) -- (becomeglideramateur)  ;
			\draw[composite] (glider) -- (gliderrule)  ;
			\draw[reference] (becomeglideramateur) -- (glider)  ;
			\draw[generalization] (rudderbeamglider) -- (rudderglider)  ;
            \draw[generalization]
            (framerudder) -- (rudder);
				\draw[generalization] (rudderglider) -- (glider);

    \end{tikzpicture}
    \caption{A sub-diagram under the ``glider block" of the Block Definition Diagram for Chapter 7 in \cite{jackman1912flying}.}
    \label{fig:provided_bdd}
\end{figure}
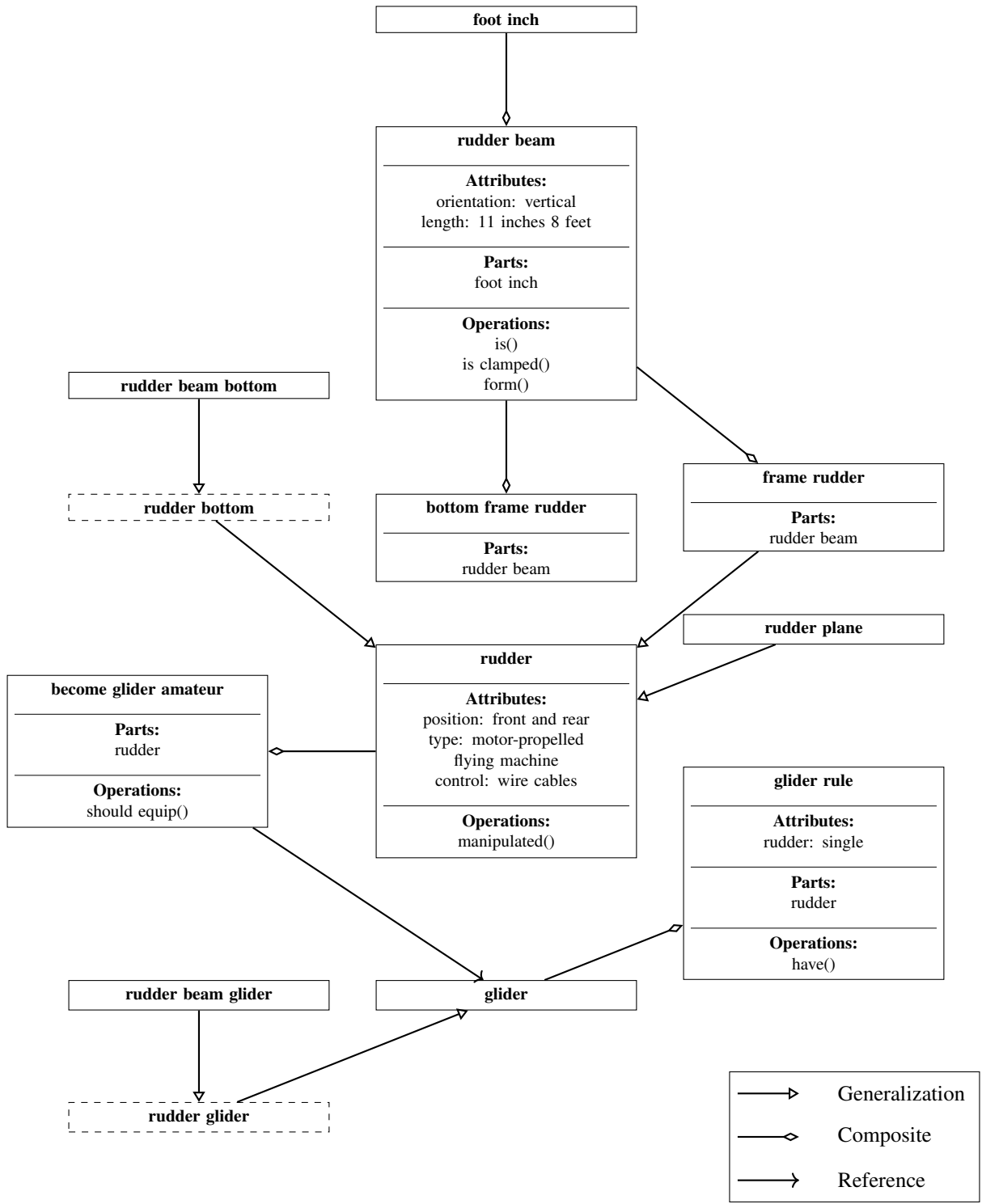

\begin{figure}
    \centering
    \begin{tikzpicture}[
        block/.style={rectangle, draw, text width=4cm, text centered, font=\footnotesize, minimum height=1.5cm},
        generalization/.style={->, thick, >={Triangle[open]}},
        composite/.style={->, thick, >={Diamond[open]}},
        reference/.style={->, thick, dashed},
        abstraction/.style={->, thick, dotted}
    ]

        \node[block] (invention) at (2, 0) {%
            \textbf{invention} \\ \rule{\linewidth}{0.3pt}
            \textbf{Attributes:} \\ 
            configuration: simple \\ 
            prevention: frequency \\ 
            weight: medium \\ 
            preferred: nice \\ 
            embodiment: present \\ 
            function: blow-by gas device \\ 
            type: embodiment \\ \rule{\linewidth}{0.3pt}
            \textbf{Parts:} \\ 
            gas device configuration, gas device \\ \rule{\linewidth}{0.3pt}
            \textbf{Operations:} \\ 
            to provide() \\ 
            is()
        };

        \node[block] (member) at (-5, 0) {%
            \textbf{member} \\ \rule{\linewidth}{0.3pt}
            \textbf{Attributes:} \\ 
            fastening: this \\ 
            turbine housing: 84a \\ 
            type: high heat portion \\ 
            method: not limited to this \\ \rule{\linewidth}{0.3pt}
            \textbf{Parts:} \\ 
            contact turbine housing \\ \rule{\linewidth}{0.3pt}
            \textbf{Operations:} \\ 
			is() \\
            having() \\ 
            fastened() \\ 
            is extended() \\ 
            is arranged() \\ 
            receives() \\ 
            may be brought()
        };

        \node[block] (turbine) at (-10, -4) {%
            \textbf{turbine} \\ \rule{\linewidth}{0.3pt}
            \textbf{Attributes:} \\ 
            type: turbine housing (hot heat portion) \\ 
			number: 84 \\
            location: exhaust path \\ \rule{\linewidth}{0.3pt}
            \textbf{Operations:} \\ 
            is provided()
        };

        \node[block] (turbine_housing) at (-5, -6.5) {%
            \textbf{turbine housing} \\ \rule{\linewidth}{0.3pt}
            \textbf{Parts:} \\ 
            member \\ \rule{\linewidth}{0.3pt}
            \textbf{Operations:} \\ 
            is()
        };

        \node[block] (turbine_catalyst_muffler) at (0, -6) {%
            \textbf{turbine catalyst muffler} \\ \rule{\linewidth}{0.3pt}
            \textbf{Operations:} \\ 
            are provided()
        };

        \node[block] (fastening_clamp_member) at (-10, 0) {%
            \textbf{fastening clamp member} \\ \rule{\linewidth}{0.3pt}
            \textbf{Operations:} \\ 
            has been described()
        };

        \node[block] (contact_turbine_housing) at (-10, 3) {%
            \textbf{contact turbine housing}
        };

		\matrix [draw, below right, xshift=-3cm, yshift=-1cm, column sep=0.5cm, row sep=0.2cm] at (turbine_catalyst_muffler.south east) {
		\draw[generalization] (0,0) -- +(1,0); & \node[anchor=west] {Generalization}; \\
		\draw[composite] (0,0) -- +(1,0); & \node[anchor=west] {Composite}; \\
		\draw[reference] (0,0) -- +(1,0); & \node[anchor=west] {Reference}; \\
	};

		\draw[composite] (member) -- (turbine_housing);
        \draw[composite] (member) -- (contact_turbine_housing);
        \draw[generalization] (turbine) -- (member);
        \draw[generalization] (turbine_housing) -- (turbine);
        \draw[generalization] (turbine_catalyst_muffler) -- (turbine_housing);
        \draw[generalization] (fastening_clamp_member) -- (contact_turbine_housing);

    \end{tikzpicture}
    \caption{A sub-diagram of the Block Definition Diagram for a patent document JP6875871B2 \cite{yasuda2021jp6875871b2} under the ``turbine" block generated using the current methodology.}
    \label{fig:patent_bdd}
\end{figure}
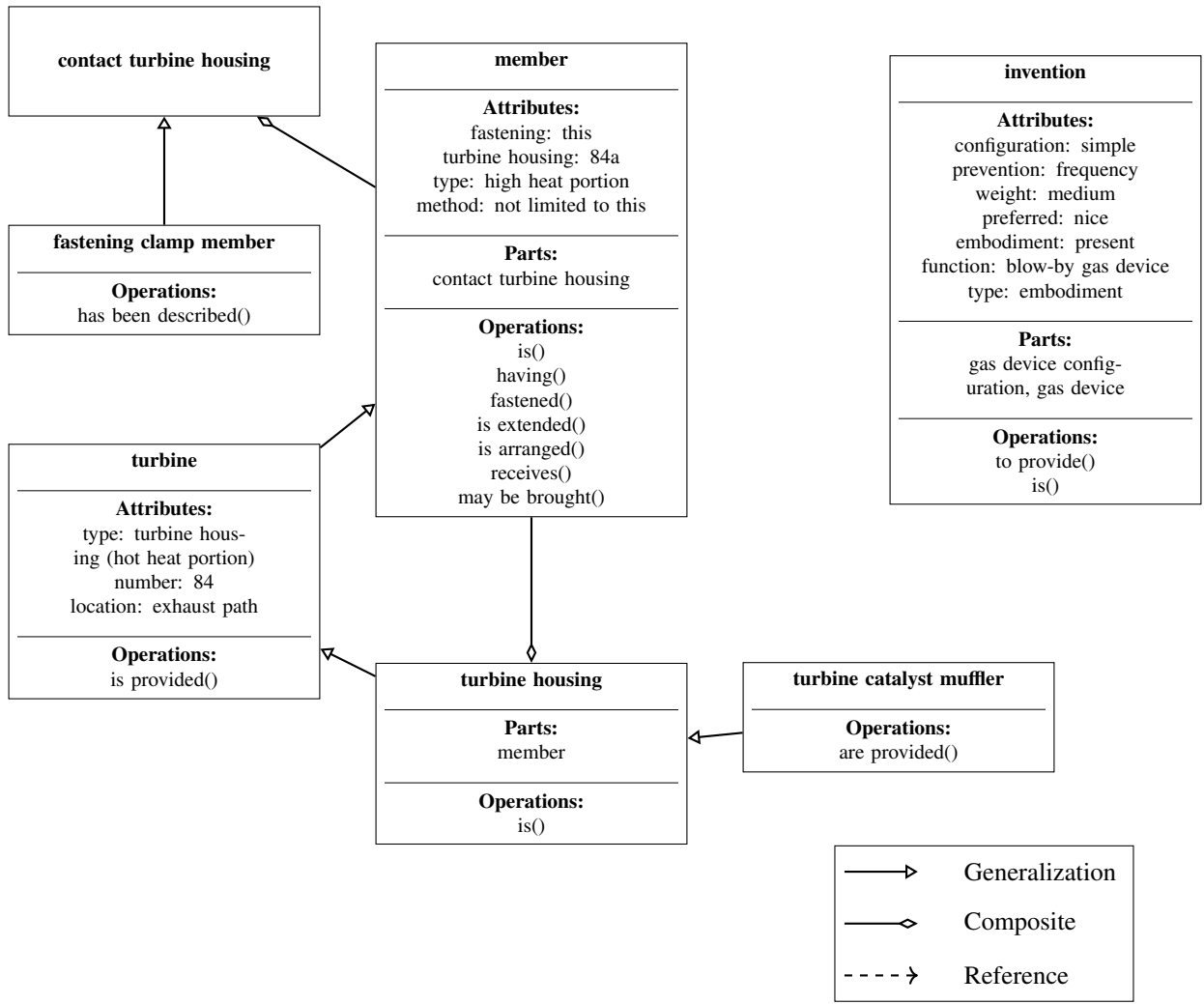

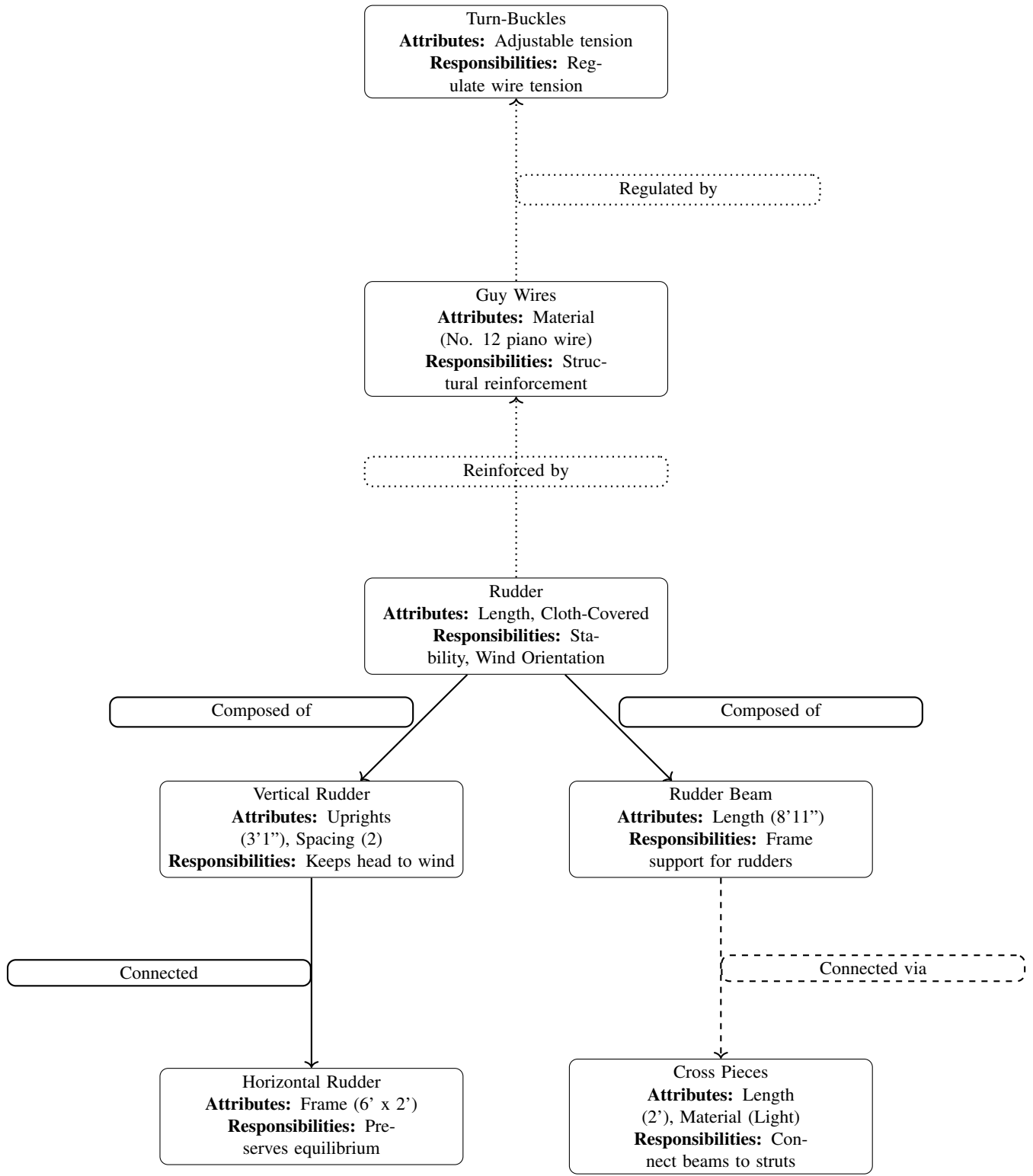
\begin{figure}
		\centering
	\begin{tikzpicture}[
		node distance=2cm and 2cm,
		every node/.style={draw, text width=5cm, align=center, font=\small, rounded corners},
		composed/.style={->, thick},
		connected/.style={->, thick, dashed},
		reinforced/.style={->, thick, dotted}
	]
	
	\tikzstyle{box} = [rectangle, draw, text width=5cm, align=center, font=\small, rounded corners]
	
	\node[box] (rudder) {Rudder \\ \textbf{Attributes:} Length, Cloth-Covered \\ \textbf{Responsibilities:} Stability, Wind Orientation};
	\node[box, below left of= rudder, node distance = 5cm] (vertical) {Vertical Rudder \\ \textbf{Attributes:} Uprights (3'1''), Spacing (2) \\ \textbf{Responsibilities:} Keeps head to wind};

	\node[box, below of= vertical, node distance = 5cm] (horizontal) {Horizontal Rudder \\ \textbf{Attributes:} Frame (6' x 2') \\ \textbf{Responsibilities:} Preserves equilibrium};

	\node[box, below right of= rudder, node distance = 5cm] (beam) {Rudder Beam \\ \textbf{Attributes:} Length (8'11'') \\ \textbf{Responsibilities:} Frame support for rudders};

	\node[box, below of= beam, node distance = 5cm] (cross) {Cross Pieces \\ \textbf{Attributes:} Length (2'), Material (Light) \\ \textbf{Responsibilities:} Connect beams to struts};

	\node[box, above of= rudder, node distance = 5cm] (guy) {Guy Wires \\ \textbf{Attributes:} Material (No. 12 piano wire) \\ \textbf{Responsibilities:} Structural reinforcement};

	\node[box, above of= guy, node distance = 5cm] (turn) {Turn-Buckles \\ \textbf{Attributes:} Adjustable tension \\ \textbf{Responsibilities:} Regulate wire tension};
	
	\draw[composed] (rudder) -- node[above left] {Composed of} (vertical);
	\draw[composed] (rudder) -- node[above right] {Composed of} (beam);
	\draw[composed] (vertical) -- node[left] {Connected} (horizontal);
	\draw[connected] (beam) -- node[right] {Connected via} (cross);
	\draw[reinforced] (rudder) -- node[above] {Reinforced by} (guy);
	\draw[reinforced] (guy) -- node[right] {Regulated by} (turn);
	
	\end{tikzpicture}
	\caption{BDD diagram generated by zero-shot prompting GPT-4o.}
	\label{fig:gpt4_bdd}
	\end{figure}
	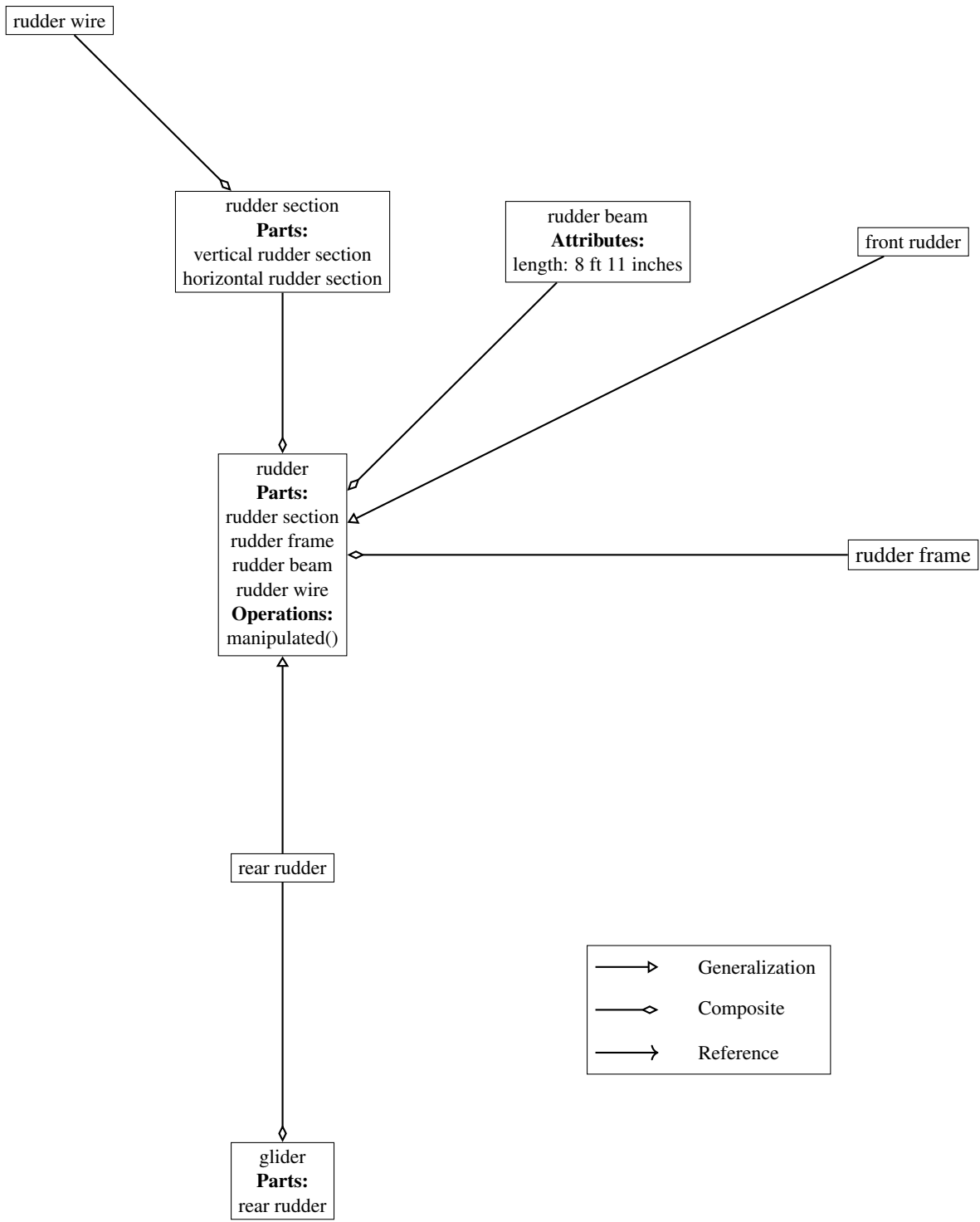
\begin{figure}
		\centering
		\begin{tikzpicture}[
			node distance=5cm,
			block/.style={draw, rectangle, align=center, font=\small},
			generalization/.style={->, thick, >={Triangle[open]}},
			composite/.style={->, thick, >={Diamond[open]}},
			reference/.style={->, thick}
		]

		\node[block] (rudder_wire) {rudder wire};
		\node[block, below right of= rudder_wire] (rudder_section) {rudder section \\ \textbf{Parts:} \\ vertical rudder section \\ horizontal rudder section};
		\node[block, right of= rudder_section] (rudder_beam) {rudder beam \\ \textbf{Attributes:} \\ length: 8 ft 11 inches};
		\node[block, below of= rudder_section] (rudder) {rudder \\ \textbf{Parts:} \\ rudder section \\ rudder frame \\ rudder beam \\
        rudder wire\\
        \textbf{Operations:} \\ manipulated()};
		\node[block, right of= rudder_beam] (front_rudder) {front rudder};
		\node[draw, below of= front_rudder] (rudder_frame) {rudder frame};
		\node[block, below of= rudder] (rear_rudder) {rear rudder};
		\node[block, below of= rear_rudder] (glider) {glider \\ \textbf{Parts:} \\ rear rudder};

		\matrix [block, below right, xshift=4cm, yshift=-1cm, column sep=0.5cm, row sep=0.2cm] at (rear_rudder.south east) {
		\draw[generalization] (0,0) -- +(1,0); & \node[anchor=west] {Generalization}; \\
		\draw[composite] (0,0) -- +(1,0); & \node[anchor=west] {Composite}; \\
		\draw[reference] (0,0) -- +(1,0); & \node[anchor=west] {Reference}; \\
	};

		\draw[composite] (rudder_wire) --  (rudder_section);
		\draw[composite] (rudder_section) --  (rudder);
		\draw[composite] (rudder_beam) -- (rudder);
		\draw[composite] (rudder_frame) --  (rudder);
		\draw[generalization] (front_rudder) --   (rudder);
		\draw[generalization] (rear_rudder) -- (rudder);
		\draw[composite] (rear_rudder) --   (glider);
	
	\end{tikzpicture}
	\caption{Ground truth BDD diagram generated manually from from Chapter 7 in \cite{jackman1912flying} under the parent block ``glider".}
	\label{fig:ground_truth_bdd}
	\end{figure}

\section{Attribute Value Extraction Prompt}

	\begin{lstlisting}[language=Python, caption={Prompt template for attribute value extraction using a Llama 3.2 3B Instruct model.}, label={lst:attribute_value_extraction_prompt}]
		prompt_template = [{
			"role": "system",
			"content": "You are a world class algorithm for value attribute extraction for engineering documents in structured formats."
			},
			{
				"role": "human",
				"content": f"""
			Extract the value attribute pairs for each entity in the given sentences, and respond with a JSON object. If an attribute is not present in the input sentence, return an empty JSON and nothing else.  Do not generate additional inputs, questions or text. The following is an example:
			Input: The engine is a Bleriot, with 5 hp and weighs 5.5kg.
			Output: {example_schema}
			"""
			},]
		example_schema = {
				"name": "engine",
				"attributes": [
					{
						"name": "power",
						"type": "int",
						"unit": "hp",
						"value": 5
					},
					{
						"name": "weight",
						"type": "float",
						"unit": "kg",
						"value": 5.5
					},
					{
						"name": "model",
						"type": "str",
						"value": "Bleriot"
					}
				]
				}
		\end{lstlisting}
	
	\section{Example Generated Code Structure}
		\begin{lstlisting}[language=Python, caption={Example generated code structure for the motion block in a simple pendulum system.}, label={lst:Motion_class_example}]
		class Motion:
			"""
			The motion is periodic and can be approximated by simple harmonic motion for small angles .
			"""

			def __init__(self):
				self.Condition = "small angles"
				self.Approximation = "simple harmonic"
				self.Angle = "small"
				self.Type = "periodic"

			def CanBeApproximated(self):
				"""
				Gravity causes the pendulum to swing .The motion is periodic and can be approximated by simple harmonic motion for small angles .The pendulum 's period depends on period length and gravity .
				"""
				pass

		\end{lstlisting}

        \section{Grid of BDD blocks for Attribute Value Extraction Validation}

        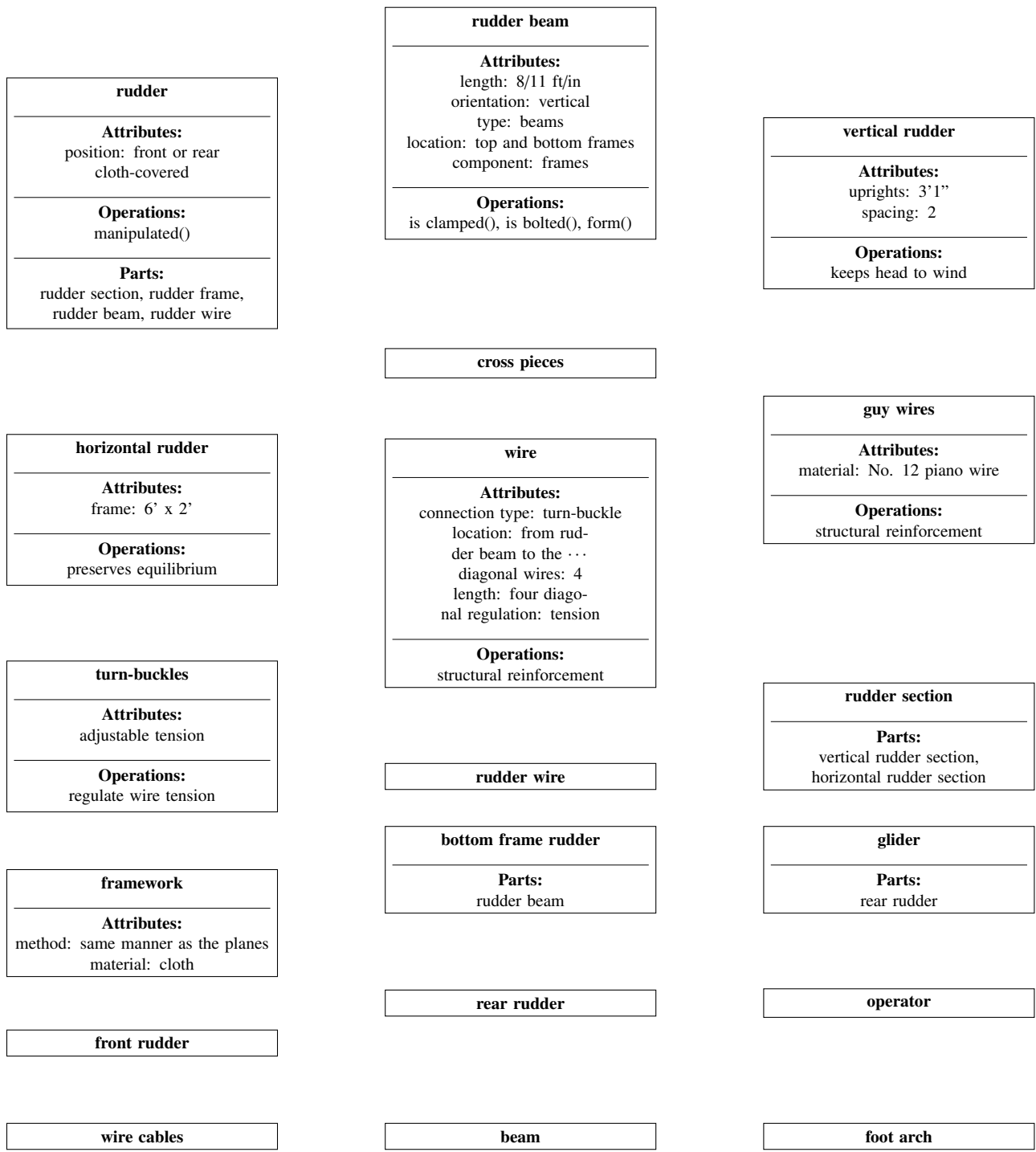
\begin{figure}
    \centering
    \begin{tikzpicture}[
		block/.style={rectangle, draw, text width=4.2cm, text centered, font=\footnotesize},
		x=6.2cm, y=-2.2cm
    ]
        \node[block] (rudder) at (0,0) {%
            \textbf{rudder} \\ \rule{\linewidth}{0.3pt}
            \textbf{Attributes:} \\
            position: front or rear \\
            cloth-covered \\
            \rule{\linewidth}{0.3pt}
            \textbf{Operations:} \\
            manipulated() \\
            \rule{\linewidth}{0.3pt}
            \textbf{Parts:} \\
            rudder section, rudder frame, rudder beam, rudder wire
        };
        \node[block] (rudderbeam) at (1,-0.6) {%
            \textbf{rudder beam} \\ \rule{\linewidth}{0.3pt}
            \textbf{Attributes:} \\
            length: 8/11 ft/in \\
            orientation: vertical \\
			type: beams \\
			location: top and bottom frames \\
			component: frames \\
            \rule{\linewidth}{0.3pt}
            \textbf{Operations:} \\
            is clamped(), is bolted(), form()
        };
        \node[block] (verticalrudder) at (2,0) {%
            \textbf{vertical rudder} \\ \rule{\linewidth}{0.3pt}
            \textbf{Attributes:} \\
            uprights: 3'1'' \\
            spacing: 2 \\
            \rule{\linewidth}{0.3pt}
            \textbf{Operations:} \\
            keeps head to wind
        };

		\node[block] (horizontalrudder) at (0,2.3) {%
			\textbf{horizontal rudder} \\ \rule{\linewidth}{0.3pt}
			\textbf{Attributes:} \\
			frame: 6' x 2' \\
			\rule{\linewidth}{0.3pt}
			\textbf{Operations:} \\
			preserves equilibrium
		};
		\node[block] (crosspiece) at (1,1.2) {%
			\textbf{cross pieces} 
		};

		\node[block] (wire) at (1,2.7) {%
			\textbf{wire} \\ \rule{\linewidth}{0.3pt}
			\textbf{Attributes:} \\
			connection type: turn-buckle \\
			location: from rudder beam to the $\cdots$ \\
			diagonal wires: 4 \\
			length: four diagonal
			regulation: tension
			\rule{\linewidth}{0.3pt}
			\textbf{Operations:} \\
			structural reinforcement
		};

		\node[block] (guywires) at (2,2) {%
			\textbf{guy wires} \\ \rule{\linewidth}{0.3pt}
			\textbf{Attributes:} \\
			material: No. 12 piano wire \\
			\rule{\linewidth}{0.3pt}
			\textbf{Operations:} \\
			structural reinforcement
		};

				\node[block] (turnbuckles) at (0,4) {%
					\textbf{turn-buckles} \\ \rule{\linewidth}{0.3pt}
					\textbf{Attributes:} \\
					adjustable tension \\
					\rule{\linewidth}{0.3pt}
					\textbf{Operations:} \\
					regulate wire tension
				};
				\node[block] (rudderwire) at (1,4.3) {%
					\textbf{rudder wire} 
				};
				\node[block] (ruddersection) at (2,4) {%
					\textbf{rudder section} \\ \rule{\linewidth}{0.3pt}
					\textbf{Parts:} \\
					vertical rudder section, horizontal rudder section
				};

				\node[block] (framework) at (0,5.4) {%
					\textbf{framework}
					\\ \rule{\linewidth}{0.3pt}
					\textbf{Attributes:} \\
					method: same manner as the planes \\
					material: cloth
				};
				\node[block] (bottomframerudder) at (1,5) {%
					\textbf{bottom frame rudder} \\ \rule{\linewidth}{0.3pt}
					\textbf{Parts:} \\
					rudder beam
				};
				\node[block] (glider) at (2,5) {%
					\textbf{glider} \\ \rule{\linewidth}{0.3pt}
					\textbf{Parts:} \\
					rear rudder
				};

				\node[block] (frontrudder) at (0,6.3) {%
					\textbf{front rudder}
				};
				\node[block] (rearrudder) at (1,6) {%
					\textbf{rear rudder}
				};
				\node[block] (operator) at (2,6) {%
					\textbf{operator}
				};

				\node[block] (wirecables) at (0,7) {%
					\textbf{wire cables}
				};
				\node[block] (beam) at (1,7) {%
					\textbf{beam}
				};
				\node[block] (footarch) at (2,7) {%
					\textbf{foot arch}
				};


    \end{tikzpicture}
    \caption{Some of the blocks of the BDD diagram generated from Chapter 7 from \cite{jackman1912flying}. The blocks are arranged in a grid format without arrows representing their relationships for better readability. There is insufficient space to show all the blocks in the BDD diagram, so only shown a subset of the blocks is shown.}
    \label{fig:bdd_attr_grid}
\end{figure}

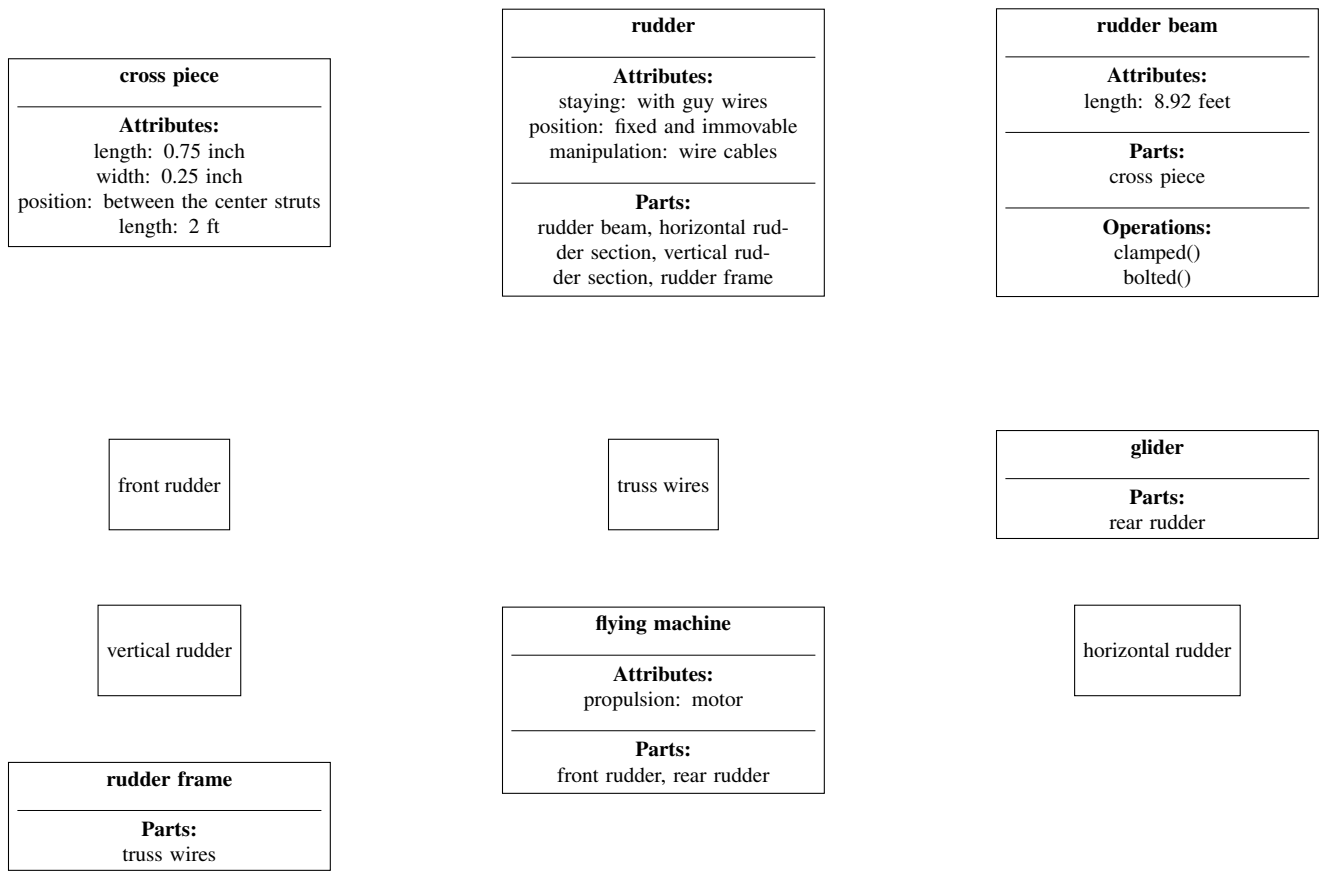
\begin{figure}
    \centering
    \begin{tikzpicture}[
        block/.style={rectangle, draw, font=\footnotesize, text width=4cm, align=center},
        plainblock/.style={rectangle, draw, font=\footnotesize, minimum height=1.2cm, align=center},
        x=6.5cm, y=-2.2cm
    ]
        \node[block] (crosspiece) at (0,0) {%
            \textbf{cross piece} \\ \rule{\linewidth}{0.3pt}
            \textbf{Attributes:} \\
            length: 0.75 inch \\
            width: 0.25 inch \\
            position: between the center struts \\
            length: 2 ft
        };
        \node[block] (rudder) at (1,0) {%
            \textbf{rudder} \\ \rule{\linewidth}{0.3pt}
            \textbf{Attributes:} \\
            staying: with guy wires \\
            position: fixed and immovable \\
            manipulation: wire cables \\
            \rule{\linewidth}{0.3pt}
            \textbf{Parts:} \\
            rudder beam, horizontal rudder section, vertical rudder section, rudder frame
        };
        \node[block] (rudderbeam) at (2,0) {%
            \textbf{rudder beam} \\ \rule{\linewidth}{0.3pt}
            \textbf{Attributes:} \\
            length: 8.92 feet \\
            \rule{\linewidth}{0.3pt}
            \textbf{Parts:} \\
            cross piece \\
            \rule{\linewidth}{0.3pt}
            \textbf{Operations:} \\
            clamped() \\
            bolted()
        };

        \node[plainblock] (frontrudder) at (0,2) {front rudder};
        \node[plainblock] (trusswires) at (1,2) {truss wires};
        \node[block] (glider) at (2,2) {%
            \textbf{glider} \\ \rule{\linewidth}{0.3pt}
            \textbf{Parts:} \\
            rear rudder
        };

        \node[plainblock] (verticalrudder) at (0,3) {vertical rudder};
        \node[block] (flyingmachine) at (1,3.3) {%
            \textbf{flying machine} \\ \rule{\linewidth}{0.3pt}
            \textbf{Attributes:} \\
            propulsion: motor \\
            \rule{\linewidth}{0.3pt}
            \textbf{Parts:} \\
            front rudder, rear rudder
        };
        \node[plainblock] (horizontalrudder) at (2,3) {horizontal rudder};
        \node[block] (rudderframe) at (0,4) {%
            \textbf{rudder frame} \\ \rule{\linewidth}{0.3pt}
            \textbf{Parts:} \\
            truss wires
        };

    \end{tikzpicture}
    \caption{Ground truth blocks of the BDD diagram generated from a document from the flying machines book, styled in a single row as in the recent example.}
    \label{fig:bdd_attr_grid_ground_truth}
\end{figure}

    \section{Dynamical System Function Templates}
    \begin{table}
	\centering
	\begin{tabular}{|p{3cm}|p{6cm}|p{3cm}|p{3cm}|}
	\hline
	\textbf{Function Name} & \textbf{Description} & \textbf{States} & \textbf{Parameters} \\
	\hline
	Simple Pendulum Oscillation & 
	A simple pendulum consists of a mass (bob) attached to a fixed point by an inextensible string. Gravity drives the pendulum to swing. $\cdots$ & 
	angle, angular\_velocity & 
	length, gravity\\
	\hline
	Mass Spring Oscillation & 
	A mass-spring system consists of a mass attached to a spring. The spring exerts a restoring force proportional to the displacement from its equilibrium position. $\cdots$ &
	displacement, velocity & 
	mass, spring\_constant, previous\_displacement, previous\_velocity, time\_step \\
	\hline
	\end{tabular}
	\caption{Summary of available dynamical system function templates.}
	\label{tab:dynamical_function_dict}
	\end{table}

		\section{Example Completed Code Class}
		\label{sec:example_code_class}
		\begin{lstlisting}[language=python, caption={Example completed code class for the motion block in a simple pendulum system. It can be seen that the appropriate dynamical equation template has been selected and that the parameters necessary to execute the function have been accurately captured.}, label={lst:example_code_class}]
	import numpy as np
	from String import String

	class Motion:
		"""
		The motion is periodic and can be approximated by simple harmonic motion for small angles .
		"""
		
		def __init__(self, initial_angle, initial_angular_velocity):
			self.Approximation = "simple harmonic"
			self.Angle = "small"
			self.Condition = "small angles"
			self.Type = "periodic"
			self.angle = initial_angle
			self.angular_velocity = initial_angular_velocity


		def CanBeApproximated(self, time_step, previous_angle, previous_angular_velocity):
			"""
			Gravity causes the pendulum to swing .The motion is periodic and can be approximated by simple harmonic motion for small angles .The pendulum 's period depends on period length and gravity .
			"""

			# System Variables
			length = String.Length  # Length of the pendulum
			g = 9.81  # Gravitational acceleration

			# Simulation Variables
			time_step = time_step
			prev_theta = previous_angle
			prev_omega = previous_angular_velocity

			alpha = -(g / length) * np.sin(prev_theta)  # Angular acceleration
			omega = prev_omega + alpha * time_step
			curr_theta = prev_theta + omega * time_step

			self.angle = curr_theta
			self.angular_velocity = omega

			return {"angle": curr_theta, "angular_velocity": omega}
		def simulate(self, time_step):
			return self.CanBeApproximated(time_step, self.angle, self.angular_velocity)
		\end{lstlisting}
	
		\section{Example Dynamical Equation Implementation template}
		\label{sec:example_dynamical_equation_template}
		\begin{lstlisting}[language=python, caption={Example dynamical equation implementation template for a simple pendulum system.}, label={lst:example_dynamical_equation_template}]
		def {{function_name}}(self, time_step, previous_angle, previous_angular_velocity):
			"""
			{{function_prompt}}
			"""

			# System Variables
			length = {{length}}  # Length of the pendulum
			g = {{gravity}}  # Gravitational acceleration

			# Simulation Variables
			time_step = time_step
			prev_theta = previous_angle
			prev_omega = previous_angular_velocity

			alpha = -(g / length) * np.sin(prev_theta)  # Angular acceleration
			omega = prev_omega + alpha * time_step
			curr_theta = prev_theta + omega * time_step

			self.angle = curr_theta
			self.angular_velocity = omega

			return {"angle": curr_theta, "angular_velocity": omega}
		\end{lstlisting}
	
	\section{Example System Class}
		\label{sec:example_system_class}
		\begin{lstlisting}[language=python, caption={Example system class for a simple pendulum system.}, label={lst:example_system_class}]
from String import String
from End import End
from Kilogram import Kilogram
from Gravity import Gravity
from Motion import Motion
import numpy as np


class System:

    def __init__(self):
        self.String = String()
        self.End = End()
        self.Kilogram = Kilogram()
        self.Gravity = Gravity()
        self.Motion = Motion()
        self.Motion = Motion(0,0)
        

    def simulate(self, args):
        """
        Simulate the system with the given arguments.
        """
        time_step = args.get('time_step', 1)
        total_time = args.get('total_time', 100)

        initial_angle = args.get('initial_angle', 0)
        initial_angular_velocity = args.get('initial_angular_velocity', 0)

        self.Motion.angle = initial_angle
        self.Motion.angular_velocity = initial_angular_velocity
        
        time = []
        angle = []
        angular_velocity = []
        
        for i in range(int(total_time / time_step)):
            
            time.append(i * time_step)
            Motion_results = self.Motion.simulate(time_step)
            angle.append(Motion_results['angle'])
            angular_velocity.append(Motion_results['angular_velocity'])

            return {
                'time': time,
                'angle': angle,
                'angular_velocity': angular_velocity,
            }
            
	\end{lstlisting}

    \section{Example System Class Template} \label{sec:example_system_class_template}
    \begin{lstlisting}[language=python, caption={Example System Class Template.}, label={lst:example_system_class_template}]
{% if top_level_components %}
{% for top_level_component in top_level_components %}
from {{top_level_component}} import {{top_level_component}}
{% endfor %}
{% endif %}
import numpy as np


class System:

    def __init__(self):
        {% if top_level_non_simulatable_components%}
            {% for top_level_non_simulatable_component in top_level_non_simulatable_components%}
        self.{{top_level_non_simulatable_component}} = {{top_level_non_simulatable_component}}()
            {% endfor %}
        {% endif %}
        {% if top_level_simulatable_components %}
            {% for top_level_simulatable_component in top_level_simulatable_components %}
        self.{{top_level_simulatable_component.name}} = {{top_level_simulatable_component.name}}({{top_level_simulatable_component.args}})
            {% endfor %}
        {% endif %}
        

    def simulate(self, args):
        """
        Simulate the system with the given arguments.
        """
        time_step = args.get('time_step', 1)
        total_time = args.get('total_time', 100)

        {% if function_states%}
        {% for state in function_states %}
        initial_{{state}} = args.get('initial_{{state}}', 0)
        {% endfor %}
        {% endif %}

        {% if top_level_simulatable_components%}
        {% for top_level_simulatable_component in top_level_simulatable_components %}
        {% for function_state in top_level_simulatable_component.function_states%}
        self.{{top_level_simulatable_component.name}}.{{function_state}} = initial_{{function_state}}
        {% endfor %}
        {% endfor %}
        {% endif %}
        
        time = []
        {% if function_states %}
        {% for state in function_states %}
        {{state}} = []
        {% endfor %}
        {% endif %}
        
        for i in range(int(total_time / time_step)):
            
            time.append(i * time_step)
            {% if top_level_simulatable_components%}
                {% for top_level_simulatable_component in top_level_simulatable_components %}
            {{top_level_simulatable_component.name}}_results = self.{{top_level_simulatable_component.name}}.simulate(time_step)
                    {% if top_level_simulatable_component.function_states %}
                        {% for function_state in top_level_simulatable_component.function_states %}
            {{function_state}}.append({{top_level_simulatable_component.name}}_results['{{function_state}}'])
                        {% endfor %}
                    {% endif %}
                {% endfor %}
            {% endif %}

            return {
                'time': time,
                {% if function_states %}
                {% for state in function_states %}
                '{{state}}': {{state}},
                {% endfor %}
                {% endif %}
            }
            
    \end{lstlisting}

    \section{Hyperparameters used in the Automated Generation of SysML diagrams}

    \begin{table}[H]
		\centering
		\begin{tabular}{|p{2.35cm}|p{2.7cm}|p{4cm}|p{4cm}|}
			\hline
			\textbf{Hyperparameter} &  \textbf{Valid Ranges} & \textbf{Description} &\textbf{Values Used in this Paper} \\
			\hline
			$\sigma_{\text{tf-idf}}$ & 0 $<\sigma_{\text{tf-idf}}<$  1 & Sets the minimum normalized tf-idf value for a noun to be a key noun &0. Most nouns found in engineering documents tend to represent significant components due to the terse writing style.\\
			\hline
			$L_{\text{phrase}}$ & $L_{\text{phrase}}>$0 & The maximum number of words allowed in a phrase in the BDD diagram & 3-5 words depending on the engineering domain. \\
			\hline
			$\sigma_{p}$ &  0 $<\sigma_{p}<$  1 & The minimum importance metric value (as defined in \Cref{sec:key_phrase_select}) required for a phrase to be chosen as a key phrase &0.6. Most phrases are chosen as key phrases to ensure exhaustive results as suggested in \cite{zhong2023natural}. \\
			\hline
			$\sigma_{\text{rel-difference}}$ &  0 $<\sigma_{\text{rel-difference}}<$  3 & The minimum importance metric  difference required for connected phrases to be mapped to composite relationships in \Cref{sec:rel_mapping}  & 0.5. Causes connected phrases with a sufficient score difference to be mapped into composite relations. \\
			\hline
			$\sigma_{\text{relationship}}$ & 0 $<\sigma_{\text{relationship}}< $ 1 & The minimum confidence value for a relationship to be considered relevant & 0.5. \\
			\hline
		\end{tabular}
		\caption{Hyperparameters used in the Automated Generation of SysML diagrams, with thier valid ranges, description and the values used in this paper. \cite{zhong2023natural}}
		\label{tab:recommended_hyperparameters}
	\end{table}

\end{document}